%% file: main.tex
\documentclass[10pt,twocolumn,letterpaper]{article}

\usepackage{cvpr}

\input{preamble}

\definecolor{cvprblue}{rgb}{0.21,0.49,0.74}
\usepackage[pagebackref,breaklinks,colorlinks,allcolors=cvprblue]{hyperref}

\usepackage[table]{xcolor}
\definecolor{ccfaBlue}{HTML}{6FA8DC}
\definecolor{ccfaGreen}{HTML}{A8D08D}
\newcommand{\best}[1]{\cellcolor{ccfaBlue!30}{#1}}
\newcommand{\second}[1]{\cellcolor{ccfaGreen!30}{#1}}

\title{RDSplat: Robust Watermarking Against Diffusion Editing for 3D Gaussian Splatting}

\author{
    Longjie Zhao$^{1}$ \quad
    Ziming Hong$^{1,*}$ \quad
    Zhenyang Ren$^{1}$ \\
    Runnan Chen$^{1,*}$ \quad
    Mingming Gong$^{2,3,*}$ \quad
    Tongliang Liu$^{1,3,*}$ \\
    \\
    $^{1}$The University of Sydney \quad
    $^{2}$The University of Melbourne \\
    $^{3}$Mohamed bin Zayed University of Artificial Intelligence \\
    \\
    \small{\texttt{\{lzha0538,zhon5578,zren0518\}@uni.sydney.edu.au\quad mingming.gong@unimelb.edu.au}} \\
    \small{\texttt{\{runnan.chen,tongliang.liu\}@sydney.edu.au}} \\
    \\
    $^{*}$Corresponding authors
}

\begin{document}

\twocolumn[{%
\maketitle
\begin{center}
\vspace{-2mm}
\includegraphics[
  width=\textwidth,
  clip,trim=5mm 98mm 35mm 0mm
]{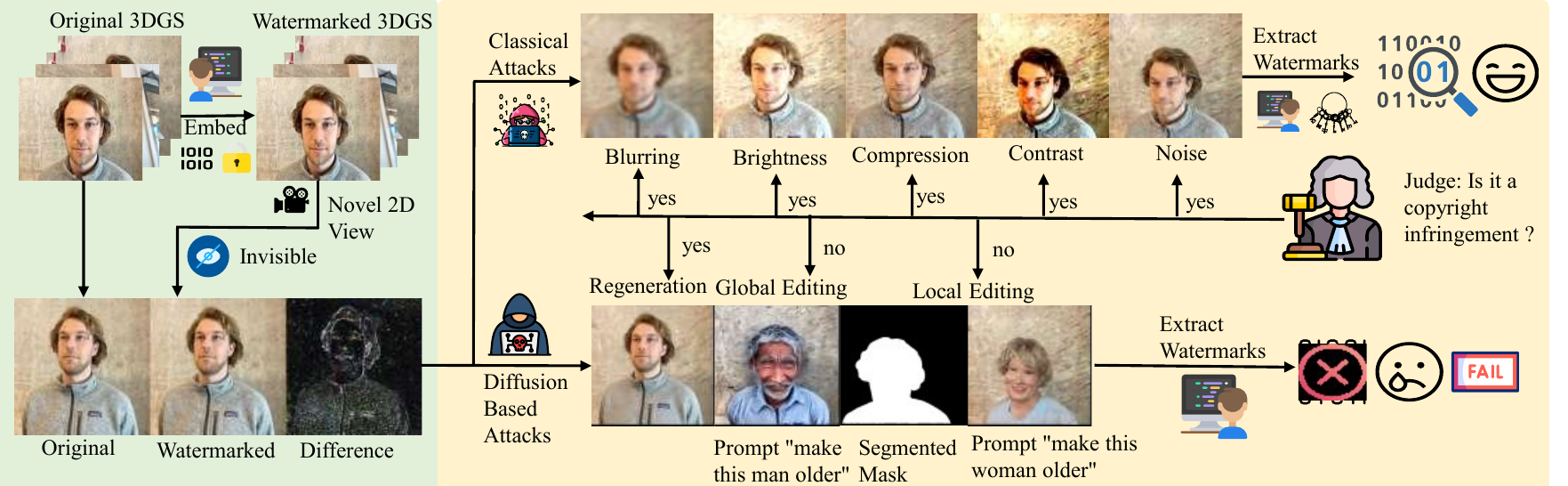}
\captionof{figure}{\textbf{Overview of 3D Watermarking and Attack Mechanisms.}
\textit{Left}: Original and watermarked 3DGS models with imperceptible differences; watermarks remain decodable from novel 2D views.
\textit{Right}: Classical attacks (blur, brightness, compression, noise) apply signal-level distortions while preserving watermark integrity. Diffusion-based editing (regeneration, global/local editing) performs semantic-level reconstruction that \emph{completely destroys} watermarks yet produces visually plausible results that appear natural to humans, enabling covert copyright infringement. A robust 3DGS watermark needs to withstand both attack categories while maintaining invisibility.}
\label{fig:motivation}
\end{center}
\vspace{2mm}
}]

\input{sec/0_abstract}

\input{sec/1_intro}

\input{sec/2_related}

\input{sec/3_preliminaries}

\input{sec/4_method}

\input{sec/5_experiments}

\input{sec/7_conclusion}

{
\small
\bibliographystyle{ieeenat_fullname}
\bibliography{main}
}

\clearpage
\appendix
\input{sec/8_suppl}

\end{document}

%% file: preamble.tex
\usepackage{amsmath}
\usepackage{amssymb}
\usepackage{algorithm}
\usepackage{algorithmic}
\usepackage{float}      
\usepackage{graphicx}   
\usepackage{caption}
\usepackage{multirow}  
\usepackage{makecell}  
\usepackage{capt-of}

\usepackage{xcolor}

%% file: sec/0_abstract.tex
\begin{abstract}
3D Gaussian Splatting (3DGS) has enabled the creation of digital assets and downstream applications, underscoring the need for robust copyright protection via digital watermarking. However, existing 3DGS watermarking methods remain highly vulnerable to diffusion-based editing, which can easily erase embedded provenance. This challenge highlights the urgent need for 3DGS watermarking techniques that are intrinsically resilient to diffusion-based editing.
In this paper, we introduce \textbf{RDSplat}, a \textbf{R}obust watermarking paradigm against \textbf{D}iffusion editing for 3D Gaussian \textbf{Splat}ting. RDSplat embeds watermarks into 3DGS components that diffusion-based editing inherently preserve, achieved through (i) proactively targeting low-frequency Gaussians and (ii) adversarial training with a diffusion proxy.
Specifically, we introduce a multi–domain framework that operates natively in 3DGS space and embeds watermarks into diffusion-editing-preserved low-frequency Gaussians via coordinated covariance regularization and 2D filtering.
In addition, we exploit the low-pass filtering behavior of diffusion-based editing by using Gaussian blur as an efficient training surrogate, enabling adversarial fine-tuning that further enhances watermark robustness against diffusion-based editing. Empirically, comprehensive quantitative and qualitative evaluations on three benchmark datasets demonstrate that RDSplat not only maintains superior robustness under diffusion-based editing, but also preserves watermark invisibility, achieving state-of-the-art performance.
\end{abstract}

%% file: sec/1_intro.tex
\section{Introduction}
\label{sec:intro}

\begin{figure}[t]
    \centering
    \includegraphics[width=\columnwidth]{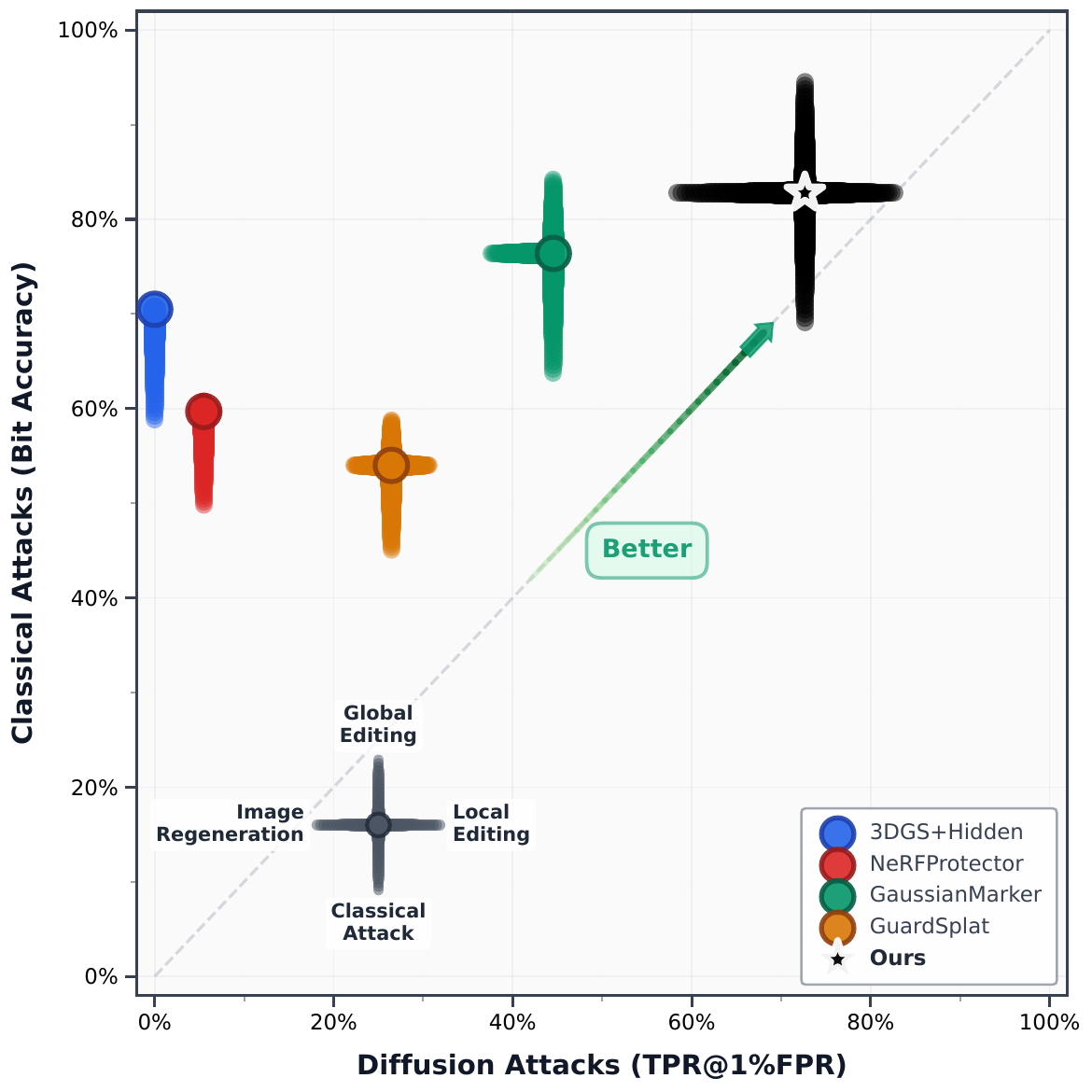}
    \caption{\textbf{Robustness comparison: classical vs. diffusion.} 
    Each method is shown as a diamond with four bars. 
    The diamond area represents encoding capacity; 
    its y-coordinate shows bit accuracy under classical attacks (averaged across multiple attack types); 
    its x-coordinate shows TPR@1\%FPR against diffusion editing. 
    Bar length indicates TPR@1\%FPR for each specific attack (longer is better). 
    Due to diffusion-based attacks' destructive nature, evaluation shifts from bit accuracy to watermark detection (TPR@1\%FPR). 
    Our method achieves balanced robustness across all attacks on the Blender dataset.}
    \vspace{-2mm}
    \label{fig:motivation2}
\end{figure}

The field of 3D representation is shifting from implicit neural fields 
(e.g., NeRF~\cite{mildenhall2021nerf}) to explicit formats, with 3D 
Gaussian Splatting (3DGS)~\cite{kerbl20233d} emerging as a prominent 
breakthrough. By modeling scenes with anisotropic Gaussians, 3DGS balances 
reconstruction fidelity and real-time rendering efficiency, driving 
adoption across dynamic scenes, autonomous driving, and content 
creation~\cite{wen20253d,zhu20243d,bao20253d,wu2024recent,fei20243d,huang2025openinsgaussian,chen2024ovgaussian,chen2024beyond,sun2025intern,chen2024panoslam,li2024urbangs,huang2025surprise3d,huang2025mllm}. As these 3DGS assets gain commercial value, protecting their copyright through digital watermarking becomes critical for ownership 
verification. Recent works have proposed 
watermarking schemes for 3DGS~\cite{huang2024gaussianmarker,chen2025guardsplat}, primarily 
targeting robustness against classical distortions such as compression, 
noise, and geometric transformations.

However, diffusion-based editing~\cite{zhu2026survey,brooks2023instructpix2pix,zhang2023adding} poses unprecedented threats to watermark robustness. Unlike classical distortions (\eg, blur, compression, noise) that apply signal-level perturbations, diffusion-based editing performs semantic-level reconstruction through iterative denoising, fundamentally altering rendered views while maintaining visual plausibility. As Fig.~\ref{fig:motivation} shows, when applied to 3DGS-rendered views, such semantic manipulations \textit{obliterate embedded watermarks, rendering existing methods ineffective}. Developing robust watermarking against both classical distortions and diffusion-based attacks becomes imperative.

Recent 2D watermarking methods (VINE~\cite{lu2024robust}, EditGuard~\cite{zhang2024editguard}) achieve diffusion robustness through frequency guidance and adversarial training. However, they cannot be directly applied to 3DGS due to architectural mismatches: 2D methods embed watermarks only in rendered images without generalization to novel views, and subtle watermark signals are blurred during 3D reconstruction from 2D watermarked images. Existing 3DGS watermarking methods employ diverse embedding strategies, operating on \emph{explicit geometric parameters}~\cite{huang2024gaussianmarker, chen2025guardsplat} or \emph{implicit feature domains}~\cite{in2025compmarkgs, li2025gaussianseal}, with some leveraging \emph{image-domain frequency cues}~\cite{jang20253d,zhang2024gs} to guide 3D embedding. However, as Fig.~\ref{fig:motivation2} illustrates, these methods exhibit severe vulnerability to diffusion-based attacks: GaussianMarker~\cite{huang2024gaussianmarker} achieves 78\% bit accuracy under classical attacks but only 45\% detection rate (TPR@1\%FPR) against diffusion-based attacks, while most baselines show weak detection under diffusion-based editing.

\begin{figure*}[t]
  \centering
  \includegraphics[width=\textwidth,trim=0 150 70 0,clip]{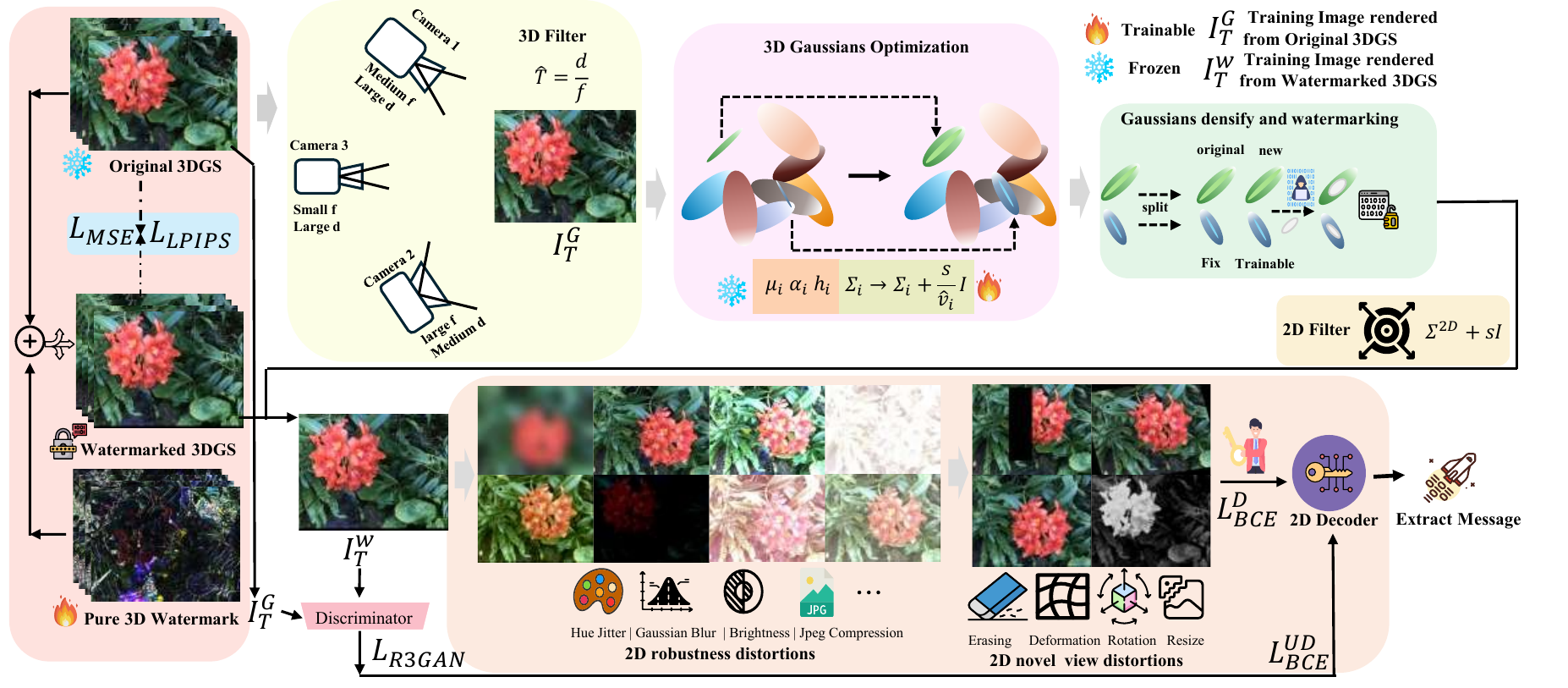}
  \caption{\textbf{Overview of the proposed RDSplat framework.}
  The proposed RDSplat model embeds robust watermarks into 3D Gaussian Splatting (3DGS) representations to protect the copyright of 3D assets against diffusion-based semantic edits.
  The pipeline consists of three main components: 
  (1) \emph{3D Gaussian optimization}, where the original 3DGS is densified and watermarked by modifying only the covariance matrices $\Sigma_i$ of the Gaussians; 
  (2) \emph{2D rendering and distortion simulation}, which generates training images under diverse robustness distortions (e.g., hue jitter, Gaussian blur, brightness, JPEG compression); and 
  (3) \emph{decoding and adversarial training}, where a 2D decoder extracts the embedded watermark while losses such as $L_{\mathrm{MSE}}$, $L_{\mathrm{LPIPS}}$, and $L_{\mathrm{BCE}}$ jointly enforce fidelity, perceptual quality, and watermark robustness.
  This joint 3D and 2D optimization ensures that the watermark remains decodable from novel 2D views even after high-level semantic or diffusion-based modifications.}
  \label{MainFramework}
\end{figure*}

Therefore, we present \textbf{RDSplat}, a novel watermarking framework 
protecting 3DGS assets against diffusion editing(Fig.~\ref{MainFramework}). \textbf{Our approach combines two complementary strategies:} \textbf{(1) Proactive low-frequency identification and embedding:} 
To leverage diffusion models' preservation of low frequencies and reconstruction of high frequencies, we develop a multi-domain 
frequency control module~\cite{shannon2006communication,nyquist2009certain,yu2024mip} that directly regularizes 3D Gaussians' 
covariance matrices in object space, complemented by 2D Mip filtering 
in screen space. This native 3D operation enables viewpoint-invariant 
low-frequency embedding, eliminating the spectral inconsistency and 
2D-to-3D back-projection artifacts. \textbf{(2) Adversarial fine-tuning with diffusion proxy:} We develop an efficient surrogate training strategy exploiting the spectral equivalence between Gaussian blur and diffusion operations, achieving two orders of magnitude speedup while maintaining robustness against actual diffusion editing through adversarial fine tuning of embedded watermarks. A pretrained decoder~\cite{zhu2018hidden,huang2024gaussianmarker} extracts watermarks for consistent verification across novel views. As Fig.~\ref{fig:motivation2} illustrates, our method uniquely occupies the best region by achieving balanced robustness with 82\% bit accuracy under classical attacks and 74\% detection rate against diffusion editing. Extensive experiments on Blender~\cite{mildenhall2021nerf}, LLFF~\cite{mildenhall2019local}, and Mip-NeRF 360~\cite{barron2022mip} demonstrate that RDSplat achieves this balanced robustness against both classical distortions and diffusion-based editing while maintaining watermark invisibility and physical consistency.

Overall, the \textbf{main contributions} of this paper are summarized as follows: \textbf{1)} We present RDSplat, a new watermarking framework that pioneers 
native 3D frequency domain control for 3DGS, achieving viewpoint-invariant 
embedding and robustness against diffusion-based editing. \textbf{2)} We present a multi-domain frequency control module that regularizes 3D Gaussian covariances in object space to control spatial frequencies, complemented by screen-space 2D Mip filtering. This native 3D approach ensures that embedded watermarks remain insensitive to diffusion-based editing. \textbf{3)} We develop an efficient surrogate training method exploiting 
the spectral equivalence between Gaussian blur and diffusion editing, reducing optimization costs. \textbf{4)} Experiments demonstrate that RDSplat achieves balanced 
robustness across both classical and diffusion attack categories, overcoming 
the tradeoff faced by prior methods.

%% file: sec/2_related.tex
\section{Related Work}
\label{sec:related}

\textbf{Robust 2D Watermarking Against Image Editing.}
Image watermarking \cite{van1994digital,zhu2018hidden,tancik2020stegastamp,wen2023tree} has long been studied for intellectual property (IP) tracking and protection \cite{fernandez2023stable,li2025rethinking,ahmadi2020redmark,guo2023domain,wang2021non,hong2024your,hong2025data,hong2024improving,hong2025toward,xiang2025jailbreaking}. Recent studies reveal that modern diffusion-based editing can covertly and severely damage watermarks~\cite{huang2025diffusion,an2024benchmarking,zhao2024invisible}.
Representative methods such as EditGuard~\cite{zhang2024editguard}, Robust-Wide~\cite{hu2024robust}, and JigMark~\cite{pan2024jigmark} explicitly train against editing pipelines using generative priors, contrastive objectives, and frequency proxies. 
VINE~\cite{lu2024robust} further shows that guiding watermarks towards lower frequencies improves survivability under editing. However, directly porting 2D strategies to 3D fails to address \emph{viewpoint-invariant decoding}: a watermark must remain decodable from novel renderings after 3D edits and re-projection, which introduces additional instability beyond 2D settings.

\noindent \textbf{3D Digital Watermarking for 3DGS.} Classical 3D watermarking explored meshes and point clouds in both spatial and spectral domains~\cite{praun1999robust,ohbuchi2002frequency}. 
The rise of 3D Gaussian Splatting (3DGS)~\cite{kerbl20233d} led to dedicated watermarking solutions~\cite{huang2024gaussianmarker,zhang2024gs,chen2025guardsplat,li2025gs,in2025compmarkgs,li2025gaussianseal}. 
3DGS has been rapidly adopted across various domains and has demonstrated impressive performance~\cite{tang2025recent,bao20253d,wu2024recent,fei20243d}. 
In contrast to Neural Radiance Fields (NeRF)~\cite{mildenhall2021nerf}, which rely on implicit neural representations for 3D scene reconstruction, 3DGS employs an explicit point-like primitive set with anisotropic covariances, enabling more interpretable and efficient rendering.

For instance, GaussianMarker~\cite{huang2024gaussianmarker} leverages uncertainty to select embedding locations and couples 3D and 2D decoders; GuardSplat~\cite{chen2025guardsplat} introduces CLIP-guided optimization and spherical harmonics (SH)-aware embedding.

Nevertheless, ensuring robustness to real-world \emph{editing attacks} in 3DGS and enabling reliable decoding directly from \emph{novel 2D views} remain largely open. 
Our work targets precisely this gap via low-frequency embedding on explicit primitives and surrogate training tailored to editing degradations.

%% file: sec/3_preliminaries.tex
\section{Preliminaries}
\noindent \textbf{3D Gaussian Splatting.} We build our framework on 3D Gaussian Splatting (3DGS)~\cite{kerbl20233d}, which represents a 3D scene as a collection of anisotropic 3D Gaussian primitives. Each Gaussian is parameterized by:
\begin{equation}
\mathcal{G} = \{\,\boldsymbol{\mu}_k,\,\Sigma_k,\,c_k,\,\alpha_k\,\}_{k=1}^{K},
\end{equation}
where $\boldsymbol{\mu}_k\!\in\!\mathbb{R}^3$ is the 3D center, $\Sigma_k\!\in\!\mathbb{R}^{3\times3}$ is the covariance matrix, $c_k$ is the color (modeled via spherical harmonics), and $\alpha_k\!\in\![0,1]$ is the opacity.

The covariance is decomposed as $\Sigma_k = R_k S_k S_k^\top R_k^\top$, where $R_k$ is a rotation matrix and $S_k$ is a scaling matrix. During rendering, each 3D Gaussian is projected to the image plane as a 2D Gaussian $G^{2\mathrm{D}}_k(\mathbf{x})$. The pixel color at $\mathbf{x}\!=\!(u,v)$ is computed via depth-ordered alpha compositing:
\begin{equation}
I(\mathbf{x}) = \sum_{k=1}^{K} c_k\,\alpha_k\,G^{2\mathrm{D}}_k(\mathbf{x}) \prod_{j<k}\!\bigl(1-\alpha_j G^{2\mathrm{D}}_j(\mathbf{x})\bigr),
\label{eq:3dgs-render}
\end{equation}
where $G^{2\mathrm{D}}_k(\mathbf{x})$ evaluates the 2D Gaussian at pixel $\mathbf{x}$, and the product term represents the transmittance from previous Gaussians.

%% file: sec/4_method.tex
\section{Method}

In this section, we introduce \textbf{RDSplat}, a novel framework for protecting 
3DGS assets against diffusion-based editing (Figure~\ref{MainFramework}). 
Our framework embeds watermarks into low-frequency regions of 3DGS minimally 
affected by diffusion-based editing via multi-domain filtering (3D object space and 
2D screen space). We employ a surrogate training strategy where Gaussian blur 
mimics diffusion-based attacks, achieving faster training. Watermark 
invisibility is maintained via R3GAN discriminator and reconstruction losses. 
A 2D decoder~\cite{zhu2018hidden} extracts the watermark, enabling robust 
decoding across diverse views after diffusion-based perturbations.

\subsection{Low Frequency Watermark Embedding via Multi-Domain Filtering}
\label{sec:lowpass}

\noindent\textbf{Motivation: Why 3D low frequency embedding?}
Unlike 3D-GSW~\cite{jang20253d}, which embeds watermarks in high-frequency regions via 2D DFT (a 2D-to-3D indirect approach), we adopt a fundamentally different strategy: \emph{embedding watermarks directly into low-frequency components of 3D Gaussian primitives through pure 3D spatial frequency analysis}. This design is motivated by observations from recent diffusion-based editing studies~\cite{lu2024robust,zhang2024editguard}:~diffusion models predominantly preserve low-frequency semantic structures while modifying high-frequency details, making low-frequency embedding naturally robust to semantic editing.

\noindent \textbf{Editing attacks destroy high frequencies.} Diffusion models~\cite{brooks2023instructpix2pix,zhang2023adding} inherently apply low-pass filtering through iterative denoising, disproportionately attenuating high frequency watermark signals. Through Fourier analysis (see Appendix~\ref{sec:freq_visualizations}), we demonstrate that diffusion editing preserves 48.53\% of low frequency energy but only 0.09\% of high-frequency energy. While all frequency bands (high, medium, and low) suffer severe degradation, low frequencies retain approximately 50\% of their energy, suggesting that embedding watermarks in low frequency regions is a viable approach. Thanks to Mip-Splatting~\cite{yu2024mip}, which treats 3D Gaussian primitives as 3D frequency filters and approximates their 2D projections using 2D Gaussian kernels, we can efficiently determine where to embed watermarks and how to control the Gaussian frequency characteristics.

\noindent \textbf{Selection of Low Frequency Gaussians.}
We identify stable low frequency primitives by computing the view dependent Nyquist frequency $\hat{\nu}_k$ for each Gaussian $G_k$. Among all cameras observing primitive $\mathbf{p}_k$, we select the maximum sampling frequency~\cite{shannon2006communication,nyquist2009certain,yu2024mip}:
\begin{equation}
\hat{\nu}_k = \max_{c \in \mathcal{V}_k} \frac{f_c}{d_c},
\label{eq:max_sampling_rate}
\end{equation}
where $f_c$ is the focal length, $d_c = \|\mathbf{p}_k - \mathbf{t}_c\|$ is the depth, and $\mathcal{V}_k$ denotes training views observing $\mathbf{p}_k$. We select Gaussians with $\hat{\nu}_k$ below the 25th percentile and visible in at least $\lceil N/3 \rceil$ training views, ensuring both frequency stability and multiview observability.

\noindent \textbf{3D Smoothing Filter for Frequency Regularization (Object Space).}
To limit the maximum frequency of each Gaussian primitive to half the sampling frequency (satisfying the Nyquist limit), we apply a 3D Gaussian low-pass filter~\cite{yu2024mip} in object space. This is achieved by convolving the original Gaussian $\mathcal{G}_k$ with covariance $\mathbf{\Sigma}_k$ with an isotropic Gaussian filter:
\begin{equation}
\mathcal{G}_k^\text{reg}(\mathbf{x}) = 
\left(\mathcal{G}_k * \mathcal{G}_\text{low}^k\right)(\mathbf{x}),
\label{eq:3d_convolution}
\end{equation}
where $\mathcal{G}_\text{low}^k$ is an isotropic Gaussian with variance $\lambda/\hat{\nu}_k^2$ and $\lambda$ is a hyperparameter controlling the filter strength. Since the convolution of two Gaussians yields another Gaussian with summed covariances, we obtain the regularized covariance:
\begin{equation}
\mathbf{\Sigma}_k^{\mathrm{reg}} = \mathbf{\Sigma}_k + \frac{\lambda}{\hat{\nu}_k^2} \mathbf{I},
\label{eq:cov_reg_method}
\end{equation}
with normalization factor $\kappa_k = \sqrt{|\mathbf{\Sigma}_k|/|\mathbf{\Sigma}_k^\text{reg}|}$. This regularization becomes an intrinsic part of the 3D representation after training, eliminating high-frequency artifacts when rendering at higher resolutions than the training scale.

\noindent \textbf{Densification and Watermark Injection.}
After identifying and regularizing high frequency Gaussians, we perform strategic densification to create dedicated watermark carriers. Our approach is motivated by a critical frequency tradeoff: directly embedding watermarks in original high frequency Gaussians $\{\mathcal{G}_k | k \in \mathcal{H}\}$ preserves visual quality and invisibility but lacks robustness against diffusion based editing, which acts as a low pass filter. Conversely, embedding in the lowest frequency Gaussians severely degrades scene appearance. 

To balance invisibility and diffusion robustness, we leverage the frequency reduced Gaussians obtained from 3D smoothing regularization. Specifically, the regularized Gaussians $\{\mathcal{G}_k^\text{reg}\}$ satisfy:
\begin{equation}
f_{\text{max}}(\mathcal{G}_k^\text{reg}) < f_{\text{max}}(\mathcal{G}_k), \quad \forall k \in \mathcal{H},
\label{eq:freq_reduction}
\end{equation}
where $f_{\text{max}}$ denotes the maximum frequency content. These regularized primitives occupy a \textbf{mid frequency band} which lower than the original high frequency components but higher than scene critical lowest frequencies:
\begin{equation}
f_{\text{min}} < f_{\text{max}}(\mathcal{G}_k^\text{reg}) < f_{\text{max}}(\mathcal{G}_k).
\label{eq:freq_hierarchy}
\end{equation}

We freeze all regularized Gaussians to preserve rendering fidelity, then densify by sampling $M$ new Gaussians from each regularized high frequency primitive's distribution:
\begin{equation}
\mathbf{p}_k^{(m)} \sim \mathcal{N}(\mathbf{p}_k, \mathbf{\Sigma}_k^{\mathrm{reg}}), \quad m=1,\ldots,M, \quad k \in \mathcal{H},
\label{eq:densification}
\end{equation}
initializing other attributes (color, opacity, rotation) from the parent Gaussian. Only these newly spawned Gaussians $\{\mathcal{G}_k^{(m)}\}$ are optimized to embed a 48 bit binary message $\mathbf{w} \in \{0,1\}^{48}$, guided by a pretrained 2D decoder $\mathcal{D}_{\theta}$~\cite{zhu2018hidden,huang2024gaussianmarker}:
\begin{equation}
\min_{\{\mathcal{G}_k^{(m)}\}} \mathcal{L}_{\text{watermark}}(\mathcal{D}_{\theta}, \mathbf{w}) + \lambda_{\text{render}} \mathcal{L}_{\text{render}}.
\label{eq:watermark_opt}
\end{equation}

This freeze then densify strategy concentrates watermark capacity in mid frequency primitives (Eq.~\ref{eq:freq_hierarchy}), achieving both visual imperceptibility and robustness to diffusion based low pass filtering, while preserving scene appearance through frozen original Gaussians. Hereafter, we refer to these regularized high frequency Gaussians as \textit{low frequency Gaussians} for brevity, emphasizing their reduced frequency relative to the original high frequency primitives.

\noindent \textbf{2D Mip Filter for Screen Space Smoothing.}
While 3D smoothing addresses aliasing during zoom in by regularizing high frequency Gaussians into low frequency ones (Eq.~\ref{eq:cov_reg_method}), aliasing can still occur during zoom out or at lower sampling rates. We augment the rasterization process with a 2D Mip filter that expands the projected 2D covariance~\cite{yu2024mip}:
\begin{equation}
\tilde{\mathbf{\Sigma}}_k = \mathbf{\Sigma}_k^{2\mathrm{D}} + \tau^2 \mathbf{I},
\label{eq:2d_mip}
\end{equation}
where $\tau \approx 0.3$ pixels approximates a single pixel footprint, and the normalization factor is $\kappa_k = \sqrt{|\mathbf{\Sigma}_k^{2D}|/|\tilde{\mathbf{\Sigma}}_k|}$. This further attenuates high frequency artifacts at rendering time on the fly during splatting, complementing the 3D frequency regularization in object space.

\subsection{Surrogate Training Strategy}
\label{sec:surrogate}

Directly incorporating diffusion based editors into training is computationally prohibitive. Through systematic Fourier analysis (Fig.~\ref{fig:diffusion_attacks}, Fig.~\ref{fig:classical_attacks}), we find that all eight tested diffusion methods (StoInv~\cite{song2020denoising}, DetInv~\cite{song2020denoising}, InstructPix2Pix~\cite{brooks2023instructpix2pix}, OmniGen~\cite{xiao2025omnigen}, IP-Adapter~\cite{ye2023ip}, DiffEdit~\cite{couairon2022diffedit}, ControlNet-Inpaint~\cite{zhang2023adding}) act as low pass filters, and crucially, Gaussian blur exhibits nearly identical frequency attenuation patterns. This motivates a surrogate training strategy using blur to achieve faster training while maintaining comparable robustness against diffusion attacks.

\noindent \textbf{Augmentation Pipeline.}
We compose frequency and geometric augmentations as $\mathcal{A} = \mathcal{G} \circ \mathcal{F}$, where $\mathcal{F}$ applies blur, JPEG, noise and rotation adjustments, while $\mathcal{G}$ introduces elastic deformation, erasing, rotation, and crop.

\noindent \textbf{Relation to VINE.}
Unlike VINE~\cite{lu2024robust}, which \emph{implicitly guides} watermarks toward low frequencies through surrogate training, we \emph{directly embed} watermarks into low frequency 3D Gaussian primitives via band limited covariance regularization (Section~\ref{sec:lowpass}). Surrogate attacks $\mathcal{A}$ enhance robustness rather than control frequency placement. We further extend to 3D by introducing geometric surrogates $\mathcal{G}$ for viewpoint handling and optimizing at the 3D primitive level for cross view decoding.

\noindent \textbf{Training Objective.}
A frozen pretrained decoder $D_\theta$ extracts 48-bit watermark $\mathbf{m}$ from augmented images via BCE loss:
\begin{equation}
    \mathcal{L}_{\text{wm}} = \text{BCE}(D_\theta(\mathcal{A}(\mathbf{I}_v)), \mathbf{m}).
    \label{eq:wm_loss}
\end{equation}
This teacher-student approach ensures viewpoint invariance: $\mathcal{G}$ trains on view variations while low-frequency primitives propagate watermarks consistently across renders.

\subsection{Training Objectives}
\label{sec:objectives}
We optimize watermark-carrying Gaussians with three losses. \textbf{Reconstruction} combines pixel and perceptual fidelity:
\begin{equation}
    \mathcal{L}_{\text{rec}} = \text{MSE}(\mathbf{I}^W, \mathbf{I}^G) + \lambda_{\text{lpips}} \cdot \text{LPIPS}(\mathbf{I}^W, \mathbf{I}^G).
    \label{eq:rec_loss}
\end{equation}
\textbf{Watermark loss} $\mathcal{L}_{\text{wm}}$ (Eq.~\ref{eq:wm_loss}) ensures decoding accuracy. \textbf{R3GAN adversarial loss} $\mathcal{L}_{\text{R3GAN}}$ with gradient penalties maintains invisibility under augmentations. The total loss is:
\begin{equation}
    \mathcal{L}_{\text{total}} = \lambda_{\text{rec}} \mathcal{L}_{\text{rec}} + \lambda_{\text{wm}} \mathcal{L}_{\text{wm}} + \lambda_{\text{R3GAN}} \mathcal{L}_{\text{R3GAN}}.
    \label{eq:total_loss}
\end{equation}

%% file: sec/5_experiments.tex
\section{Experiments}
\label{sec:experiments}

\begin{table}[t]
\scriptsize
\centering
\setlength{\tabcolsep}{4pt}
\renewcommand{\arraystretch}{1.2}
\caption{Image quality metrics on Blender and LLFF datasets (\textbf{test views}). The \textbf{Clean} column shows bit accuracy (\%) for watermark extraction from rendered 2D images without any distortion or attack. Complete results including train, interpolate, and all views are provided in Appendix~\ref{app:full_results}. A light blue background indicates the best performance and a light green background indicates the second best in each column.}
\label{tab:image_quality}
\begin{tabular}{lccccc}
\toprule
\textbf{Method} & \textbf{PSNR}$\uparrow$ & \textbf{SSIM}$\uparrow$ & \textbf{MSE}$\downarrow$ & \textbf{LPIPS}$\downarrow$ & \textbf{Clean}$\uparrow$ \\
\midrule
3DGS+Hidden~\cite{zhu2018hidden}      & 20.73 & 0.649 & 674.2 & 0.190 & 0.782 \\
3DGS+Vine~\cite{lu2024robust}         & \second{25.85} & 0.860 & \second{218.0} & 0.158 & 0.450 \\
NeRFProtector~\cite{luo2023copyrnerf} & 21.21 & 0.747 & 1293 & 0.222 & 0.676 \\
GaussianMarker~\cite{huang2024gaussianmarker}   & \best{28.68} & \second{0.906} & \best{142.6} & \second{0.079} & \second{0.958} \\
GuardSplat~\cite{chen2025guardsplat}       & -- & \best{0.991} & \best{7.223} & \best{0.008} & 0.602 \\
\midrule
\textbf{Ours}    & 20.52 & 0.681 & 738.1 & 0.278 & \best{0.970} \\
\bottomrule
\end{tabular}
\end{table}

\begin{table*}[t]
\scriptsize
\centering
\setlength{\tabcolsep}{3pt}
\renewcommand{\arraystretch}{1.2}
\caption{Robustness against classical attacks on Blender and LLFF datasets (test views). Values show bit accuracy (TPR@1\%FPR) averaged across 5 intensity levels (L1--L5, ranging from weak to strong). Parameters shown below each attack type correspond to Level 3 (medium intensity). Complete results are provided in Appendix~\ref{app:full_results}. \textit{3DGS+Vine shows constant 0.45 accuracy due to decoder failure on 3DGS rendered views.} A light blue background indicates the best performance and a light green background indicates the second best for each column.}
\label{tab:classical_attack}
\begin{tabular*}{\textwidth}{@{\extracolsep{\fill}}lcccccccccc@{}}
\toprule
\textbf{Method} & \textbf{Blurring} & \textbf{Brightness} & \textbf{Compression} & \textbf{Contrast} & \textbf{Erasing} & \textbf{Noise} & \textbf{Resized Crop} & \textbf{Rotation} & \textbf{Regen. VAE} & \textbf{Average} \\
& (r=10) & (1.5$\times$) & (Q=50) & (1.5$\times$) & (12.5\%) & ($\sigma$=0.05) & (0.75) & (22.5°) & & \\
\midrule
3DGS+Hidden~\cite{zhu2018hidden}    & 0.604 (0.000) & 0.776 (0.027) & 0.724 (0.000) & 0.784 (0.032) & 0.757 (0.041) & \second{0.642 (0.014)} & 0.740 (0.000) & 0.634 (0.002) & 0.687 (0.000) & 0.705 (0.013) \\
3DGS+Vine~\cite{lu2024robust}       & 0.450 (1.000) & 0.450 (1.000) & 0.450 (1.000) & 0.450 (1.000) & 0.450 (1.000) & 0.450 (1.000) & 0.450 (1.000) & 0.450 (1.000) & 0.450 (1.000) & 0.450 (1.000) \\
NeRFProtector~\cite{luo2023copyrnerf}   & 0.495 (0.000) & 0.635 (0.393) & 0.561 (0.129) & 0.640 (0.418) & 0.639 (0.394) & 0.609 (0.285) & 0.638 (0.398) & 0.560 (0.113) & -- & 0.597 (0.266) \\
GaussianMarker~\cite{huang2024gaussianmarker}  & \second{0.602 (0.371)} & \second{0.914 (0.976)} & \second{0.733 (0.826)} & \second{0.926 (0.993)} & \second{0.921 (0.970)} & 0.612 (0.232) & \second{0.831 (0.974)} & \second{0.664 (0.634)} & \second{0.671 (0.650)} & \second{0.764 (0.736)} \\
GuardSplat~\cite{chen2025guardsplat}      & 0.450 (0.046) & 0.565 (0.370) & 0.554 (0.305) & 0.578 (0.442) & 0.525 (0.252) & 0.574 (0.433) & 0.551 (0.314) & 0.521 (0.202) & 0.544 (0.252) & 0.540 (0.291) \\
\midrule
\textbf{Ours}   & \best{0.604 (0.306)} & \best{0.933 (0.999)} & \best{0.870 (0.983)} & \best{0.944 (0.999)} & \best{0.945 (0.990)} & \best{0.648 (0.370)} & \best{0.877 (0.992)} & \best{0.791 (0.941)} & \best{0.838 (0.969)} & \best{0.828 (0.840)} \\
\bottomrule
\end{tabular*}
\end{table*}

\begin{table*}[t]
\scriptsize
\centering
\setlength{\tabcolsep}{3pt}
\renewcommand{\arraystretch}{1.25}
\caption{Robustness against diffusion attacks on Blender and LLFF datasets (test views). Values are shown as bit accuracy (TPR@1\%FPR). A light blue background indicates the best performance and a light green background indicates the second-best performance in each column. \textit{3DGS+Vine shows constant 0.45 accuracy due to decoder failure on 3DGS rendered views.} Complete results are provided in Appendix~\ref{app:full_results}.}
\label{tab:diffusion_attack}
\begin{tabular*}{\textwidth}{@{\extracolsep{\fill}}lcccccccc@{}}
\toprule
\textbf{Method} & \multicolumn{2}{c}{\textbf{Image Regeneration}} & \multicolumn{4}{c}{\textbf{Global Editing}} & \textbf{Local Editing} & \textbf{Average} \\
\cmidrule(lr){2-3} \cmidrule(lr){4-7} \cmidrule(lr){8-8}
& \textbf{Deterministic} & \textbf{Stochastic} & \textbf{InstructPix2Pix} & \textbf{DiffEdit} & \textbf{IPAdapter} & \textbf{Omnigen} & \textbf{ControlNet Inpainting} & \\
\midrule
3DGS+Hidden~\cite{zhu2018hidden}        & \second{0.589 (0.000)} & 0.573 (0.000) & \second{0.598 (0.000)} & 0.579 (0.000) & 0.575 (0.000) & 0.588 (0.000) & \second{0.576 (0.000)} & 0.583 (0.000) \\
3DGS+Vine~\cite{lu2024robust}          & 0.450 (1.000) & 0.450 (1.000) & 0.450 (1.000) & 0.450 (1.000) & 0.450 (1.000) & 0.450 (1.000) & 0.450 (1.000) & 0.450 (1.000) \\
NeRFProtector~\cite{luo2023copyrnerf}      & 0.512 (0.063) & 0.433 (0.014) & -- & 0.511 (0.084) & 0.436 (0.046) & -- & 0.518 (0.061) & 0.482 (0.054) \\
GaussianMarker~\cite{huang2024gaussianmarker}     & 0.587 (0.337) & \second{0.599 (0.499)} & -- & \second{0.600 (0.478)} & \second{0.581 (0.255)} & \second{0.609 (0.562)} & -- & \second{0.599 (0.455)} \\
GuardSplat~\cite{chen2025guardsplat}         & 0.492 (0.163) & 0.533 (0.341) & 0.473 (0.188) & 0.525 (0.353) & 0.528 (0.324) & -- & 0.513 (0.253) & 0.511 (0.270) \\
\midrule
\textbf{Ours}      & \best{0.678 (0.836)} & \best{0.661 (0.892)} & \best{0.616 (0.596)} & \best{0.642 (0.800)} & \best{0.664 (0.836)} & \best{0.639 (0.702)} & \best{0.613 (0.606)} & \best{0.657 (0.739)} \\
\bottomrule
\end{tabular*}
\end{table*}

\begin{table*}[t]
\centering
\caption{Quantitative comparison in Blender and LLFF datasets. A light \textcolor{black}{blue} background indicates the best performance and a light \textcolor{black}{green} background indicates the second-best performance in each column.}
\label{tab:vine_comparison}
\resizebox{0.95\textwidth}{!}{
\begin{tabular}{l|ccccc|ccc}
\toprule
\multirow{2}{*}{Method} & 
\multicolumn{5}{c|}{Classical Editing} & 
\multicolumn{3}{c}{Diffusion Editing} \\
\cmidrule(lr){2-6} \cmidrule(lr){7-9}
& Brightness & Contrast & Erasing & Resized Crop & Rotation & Deterministic & Stochastic & IPAdapter \\
\midrule
Vine & \best{0.99 (1.00)} & \best{1.00 (1.00)} & \best{1.00 (1.00)} & \second{0.50 (0.04)} & \second{0.51 (0.03)} & \best{0.79 (0.92)} & \best{0.83 (0.95)} & \best{0.91 (1.00)} \\
Ours & \second{0.98 (1.00)} & \second{0.99 (1.00)} & \second{0.98 (1.00)} & \best{0.91 (1.00)} & \best{0.81 (0.96)} & \second{0.70 (0.87)} & \second{0.67 (0.85)} & \second{0.68 (0.83)} \\
\bottomrule
\end{tabular}
}
\end{table*}

\begin{table*}[t]
\centering
\caption{Ablation study in the LLFF dataset.}
\label{tab:main_results}
\resizebox{0.85\textwidth}{!}{
\begin{tabular}{l|cccc|ccc}
\toprule
\multirow{2}{*}{\textbf{Configuration}} & 
\multicolumn{4}{c|}{\textbf{Image Quality}} & 
\multicolumn{3}{c}{\textbf{Diffusion Attack}} \\
\cmidrule(lr){2-5} \cmidrule(lr){6-8}
& PSNR$\uparrow$ & SSIM$\uparrow$ & MSE$\downarrow$ & LPIPS$\downarrow$ & 
Deterministic & InstructPix2Pix & DiffEdit \\
\midrule
w/o data augmentation & \best{25.73} & \best{0.848} & \best{214.1} & \best{0.096} & 
0.565 (0.291) & 0.579 (0.329) & 0.559 (0.232) \\
w/o 3D f + w/ H & 16.11 & 0.551 & 1674.0 & 0.460 & 
0.652 (0.768) & \best{0.657 (0.849)} & - \\
w/o 3D f + w/ L & 20.76 & 0.711 & 698.5 & 0.276 & 
0.607 (0.467) & 0.627 (0.653) & \second{0.618 (0.667)} \\
w/o 3D f + w/ A & 18.39 & 0.526 & 986.5 & 0.413 & 
0.661 (0.823) & 0.633 (0.744) & - \\
\midrule
\textbf{Ours} & \second{22.67} & \second{0.798} & \second{432.1} & \second{0.191} & 
\best{0.674 (0.899)} & \second{0.637 (0.713)} & \best{0.648 (0.776)} \\
\bottomrule
\end{tabular}
}
\end{table*}

\begin{figure*}[t]
    \centering
    \includegraphics[width=\linewidth, trim=0 15pt 0 0, clip]{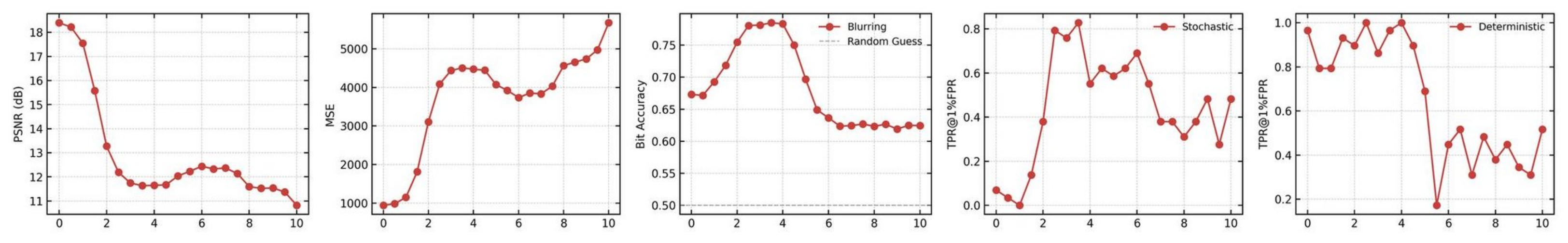}
    \caption{\textbf{Sensitivity analysis of Gaussian blur sigma during training.} Performance metrics across different blur sigma values ($\sigma$) on the LLFF flower dataset. The x-axis represents the Gaussian blur sigma size used during training. Left two plots show rendering quality metrics (PSNR, MSE), while right three plots demonstrate robustness against blur editing and diffusion editing.}
    \label{fig:blur_sensitivity2}
\end{figure*}

\subsection{Experimental Settings}

\noindent \textbf{Datasets.}
We use two popular benchmark datasets for evaluation: 
\textbf{Blender}~\cite{mildenhall2021nerf},
\textbf{LLFF}~\cite{mildenhall2019local}. Additionally, we generate interpolated viewpoints by linearly interpolating camera parameters between every pair of consecutive training camera poses in 3D space (see Appendix~\ref{subsec:interpolate_views} for detailed formulation).
All testing viewpoints are used to compute averages.
We also report metrics on \emph{train/test/interpolate} views to expose viewpoint generalization, where interpolate views represent more challenging perspectives with greater viewpoint distortion.
Detailed characteristics of each dataset are provided in Table~\ref{tab:dataset_comparison}.

\noindent \textbf{Baselines.}
We compare against: 
\textbf{(1) 3DGS native methods} that embed watermarks into 3D Gaussian primitives: GaussianMarker~\cite{huang2024gaussianmarker} and GuardSplat~\cite{chen2025guardsplat};
\textbf{(2) Adapted 2D methods} applied to 3DGS by first embedding watermarks into all input images using a 2D encoder, then training 3DGS on these watermarked images, and finally extracting watermarks from rendered views using a 2D decoder: HiDDeN~\cite{zhu2018hidden} and VINE~\cite{lu2024robust} (denoted as \textit{3DGS+Hidden} and \textit{3DGS+Vine});
\textbf{(3) NeRF based methods} for reference: NeRFProtector~\cite{luo2023copyrnerf} and NerfSignature~\cite{luo2025nerf} though architecturally incompatible due to implicit vs. explicit representation differences. NeRFSignature~\cite{luo2025nerf} is unique in that while watermarks are globally embedded in the 3D scene, they are only decodable from key-specific viewpoints, not arbitrary novel views. Extraction requires particular pose-patch combinations; thus, evaluation is conducted on training viewpoints (full results in Appendix~\ref{app:full_results}).

\subsection{Evaluation Protocol}

Following the standard protocol in 3D watermarking literature, we evaluate all methods on train, test, and interpolate views to measure both memorization and generalization capabilities. Our evaluation encompasses five key aspects: \textbf{capacity} (detailed comparisons in Appendix~\ref{app:bit_capacity}), \textbf{image quality \& invisibility} (Table~\ref{tab:image_quality}), \textbf{classical robustness} (Table~\ref{tab:classical_attack}), \textbf{editing robustness} (Table~\ref{tab:diffusion_attack}), and \textbf{time efficiency} (detailed comparisons in Appendix~\ref{app:time_efficiency}). In this section, we present results on test views. The complete results including train, interpolate, and all views are provided in Appendix~\ref{app:full_results}.

\noindent \textbf{Image Quality and Invisibility.}
Table~\ref{tab:image_quality} evaluates visual quality and watermark invisibility using PSNR$\uparrow$, SSIM$\uparrow$, MSE$\downarrow$, and LPIPS~\cite{zhang2018unreasonable}$\downarrow$. Our method achieves PSNR of 20.52 and SSIM of 0.681 on test views, balancing quality and robustness. There exists a tradeoff between image quality and robustness~\cite{li2025gaussmarker}: stronger watermark signals enhance robustness but degrade visual quality. The \textbf{Clean} column shows watermark extraction accuracy without any attacks: our method achieves 97.0\% bit accuracy, demonstrating reliable watermark embedding with strong signal strength. GuardSplat attains near-perfect invisibility (SSIM: 0.991) but exhibits lower extraction accuracy (60.2\%) in test views compared to train views (95.9\%) in Appendix~\ref{tab:classical_attack_full}, demonstrating poor generalization to novel views. GaussianMarker achieves the best visual quality (PSNR: 28.68, SSIM: 0.906) and high clean accuracy (95.8\%), though its robustness to editing attacks is limited, as shown in subsequent analysis.

\noindent \textbf{Classical Robustness.}
Table~\ref{tab:classical_attack} evaluates robustness against nine common image transformations. Our method achieves the highest average bit accuracy of \textbf{0.828} on test views, significantly outperforming all baselines. Strong robustness is maintained across photometric transformations (brightness, contrast), geometric distortions (rotation, erasing, resized crop), and compression attacks. High TPR@1\%FPR values (e.g., 0.999 for brightness/contrast, 0.983 for compression) confirm reliable detection with minimal false positives, demonstrating superior robustness compared to competing methods.

\noindent \textbf{Editing Robustness.}
Table~\ref{tab:diffusion_attack} evaluates robustness against modern diffusion based editing attacks, which pose substantially stronger threats than classical transformations. We assess seven challenging scenarios including deterministic and stochastic image regeneration, instruction guided editing (InstructPix2Pix, DiffEdit, Omnigen), adapter based manipulation (IP-Adapter), and local inpainting (ControlNet). Our method exhibits outstanding resilience, achieving an average bit accuracy of \textbf{0.657}—significantly surpassing all baselines (next best: GaussianMarker 0.599). Performance remains particularly strong under regeneration attacks (0.678 deterministic, 0.661 stochastic), where competing approaches degrade substantially. This superior editing robustness validates our low frequency embedding strategy and surrogate based training, which explicitly optimizes against diffusion model distortions. We conduct subjective experiments to evaluate watermark robustness against diffusion based editing attacks. Detailed visual comparisons are provided in Appendix~\ref{appendix:subjective}.

\noindent \textbf{Sensitivity Analysis.} As shown in Fig.~\ref{fig:blur_sensitivity2}, the training blur strength significantly impacts the trade-off between image quality and decoding robustness. Setting Gaussian blur to $\sigma \in [2, 4]$ achieves optimal performance, maintaining high image fidelity while ensuring stable decoding under both blur and diffusion attacks. However, excessively large $\sigma$ values degrade the fine-grained distribution of the embedded signal. Although higher blur strength appears to enhance blur tolerance, it actually disrupts the hidden structure, preventing the decoder from learning consistent mappings and thereby increasing decoding failure rates. Therefore, we recommend $\sigma \in [2, 4]$ to balance imperceptibility and robustness while avoiding signal degradation from over blurring. Further sensitivity analyses can be found in Appendix~\ref{appendix:AdditionalAnalysis}.

\subsection{Extensive Study in Vine and Gaussctrl}

\noindent \textbf{Comparison with VINE.} 
To evaluate the gap between native 3D watermarking and SOTA 2D diffusion-robust methods, we compare with VINE~\cite{lu2024robust}. Although VINE and RDSplat differ fundamentally in embedding strategies (2D image space vs. 3D Gaussian primitives), both extract watermarks from 2D rendered views using 2D decoders, enabling a coarse but meaningful comparison of decoding robustness. We initially attempted the 3DGS+VINE baseline (embedding watermarks into training images then training 3DGS), but it failed completely. Therefore, we directly render views from the original 3DGS, apply VINE's encoder to embed watermarks into these rendered 2D images, then extract using VINE's decoder under identical attack settings, isolating the comparison to decoding robustness on 2D views. 

Table~\ref{tab:vine_comparison} shows that RDSplat achieves comparable classical robustness with superior performance on geometric transformations: resized crop (0.91 vs. 0.50) and rotation (0.81 vs. 0.51). This advantage stems from our 3D-native embedding, which provides inherent multi-view consistency, whereas the 2D watermarks from VINE suffer from view-dependent corruption. On deterministic diffusion attacks, RDSplat attains 0.70 average accuracy compared to 0.79 for VINE. Bridging this gap would require training native 3D encoder-decoder models on a large scale 3D dataset with diffusion-aware optimization, which we leave for future work.

\noindent \textbf{Extensive study in Gaussctrl.} We conduct an extensive study using the 3D editing method GaussCtrl~\cite{wu2024gaussctrl}, as detailed in Appendix~\ref{appendix:Gaussctrl}. The pipeline works as follows: (1) embed watermarks into 3DGS using our method, (2) render to 2D views, (3) apply GaussCtrl to reconstruct an edited 3D scene, and (4) render to 2D views for watermark extraction. This process involves multiple dimensional transitions: 3D$\rightarrow$2D$\rightarrow$3D$\rightarrow$2D, demonstrating watermark robustness through repeated projection and reconstruction.

\subsection{Ablation Study}
\label{sec:ablation}
We conduct ablation studies to validate multi-domain filter and surrogate training strategy(Table~\ref{tab:main_results}). 

\noindent\textbf{Data Augmentation and surrogate training strategy.} Without augmentation, the model achieves higher image quality (PSNR: 25.73) because there is no blurring attack that would soften the Gaussians and degrade image quality. However, it shows weak performance after diffusion-based editing. Detection accuracy against stochastic attacks plummets to 0.291 vs 0.899 in ours. This validates our surrogate training strategy (Section~\ref{sec:surrogate}), which exploits the spectral equivalence between Gaussian blur and diffusion's low-pass filtering behavior. Our full model sacrifices only 3.06 PSNR but gains
significantly improved robustness (0.739 vs 0.422), demonstrating efficient adversarial fine-tuning against diffusion editing.

\noindent\textbf{3D Filter and Frequency Selection.} We compare three embedding strategies without multi-domain frequency control: high frequency (w/ H), low frequency (w/ L), and all frequencies (w/ A). High frequency embedding (w/ H) achieves the worst visual quality (PSNR: 16.11) because unfiltered high-frequency Gaussians violate the Nyquist limit, causing severe rendering artifacts. Direct low frequency embedding (w/ L) without 3D regularization produces noise and low bit accuracy, as original low-frequency primitives lack sufficient representation capacity. Our 3D covariance-based filtering (Eq.~\ref{eq:cov_reg_method}) regularizes high-frequency Gaussians into band-limited primitives below the Nyquist frequency, creating stable carriers that survive diffusion editing's low-pass filtering and maintain rendering quality via coordinated 2D Mip filtering (Eq.~\ref{eq:2d_mip}) in screen space. This native 3D frequency control achieves low-frequency embedding that targets regions naturally preserved by diffusion models.

%% file: sec/7_conclusion.tex
\section{Conclusion}
\label{sec:conclusion}

We present \textbf{RDSplat}, a robust watermarking framework for 3D Gaussian Splatting that withstands diffusion based editing attacks. Our key contributions are: (1) multi domain low frequency embedding via 3D covariance filtering and 2D Mip filtering, ensuring watermarks reside in frequency bands preserved by diffusion editors; (2) blur as surrogate training strategy, achieving faster training by exploiting the spectral similarity between Gaussian blur and diffusion editing. Extensive experiments demonstrate SOTA robustness against both classical and diffusion attacks while maintaining rendering quality and watermark invisibility.

\noindent \textbf{Future Work.}
Future directions include exploring direct 3D decoders for view agnostic extraction and unified robustness frameworks across spatial and latent domains.

%% file: sec/8_suppl.tex
\clearpage
\appendix
\setcounter{page}{1}
\setcounter{table}{0}      
\setcounter{figure}{0}     
\setcounter{section}{0}   
\renewcommand{\thetable}{A.\arabic{table}}   
\renewcommand{\thefigure}{A.\arabic{figure}} 
\maketitlesupplementary

\section{Experiments}
\subsection{Datasets}
This is detailed datasets description.
\begin{table}[h!]  
\scriptsize
\centering
\setlength{\tabcolsep}{3pt}
\renewcommand{\arraystretch}{1.2}
\caption{Comparison of datasets used for 3D scene reconstruction and watermarking.}
\label{tab:dataset_comparison}
\begin{tabular}{lcccc}
\toprule
\textbf{Dataset} & \textbf{Type} & \textbf{Real} & \makecell{\textbf{Primary}\\\textbf{Task}} & \textbf{Scenes} \\
\midrule
Blender & Indoor (Synth) & No  & Novel View Synth. & 8 \\
LLFF   & Indoor (Near)   & Yes & Robustness Testing & 9 \\

\bottomrule
\end{tabular}
\end{table}

\subsection{Compared methods}
As shown in Table~\ref{tab:method_comparison}, we present a comparative overview of representative watermarking methods categorized by their embedding frameworks, organized into three distinct technical approaches: (a) 3DGS + 2D Encoder–Decoder Watermarking; (b) NeRF-based 3D Watermarking; (c) 3DGS-based 3D Watermarking.

\begin{table}[t]
\scriptsize
\centering
\setlength{\tabcolsep}{6pt}
\renewcommand{\arraystretch}{1.2}
\caption{Representative and compared watermarking methods categorized by their embedding framework.}
\label{tab:method_comparison}
\begin{tabular}{lccc}
\toprule
\textbf{Method} & \textbf{Year} & \textbf{Venue} & \textbf{Bit Numbers} \\
\midrule
3DGS+HiDDeN~\cite{zhu2018hidden}      & 2018 & ECCV & 48 \\
3DGS+VINE~\cite{lu2024robust}        & 2025 & ICLR & 100 \\
\midrule
NeRFProtector~\cite{luo2023copyrnerf}    & 2024 & ECCV & 48 \\
NeRF Signature~\cite{luo2025nerf}   & 2025 & TPAMI & 32 \\
\midrule
GaussianMarker~\cite{huang2024gaussianmarker}   & 2024 & NeurIPS & 48 \\
GuardSplat~\cite{chen2025guardsplat}       & 2025 & CVPR & 48 \\
\bottomrule
\end{tabular}
\vspace{2mm}
\scriptsize
\end{table}

\subsection{Interpolated Views Generation}
\label{subsec:interpolate_views}

To comprehensively evaluate viewpoint generalization, we introduce \textbf{interpolated views} as an additional test set beyond the standard train/test split.

\paragraph{Method.}
Given $N$ training cameras $\{\mathcal{C}_1, \mathcal{C}_2, \ldots, \mathcal{C}_N\}$, we generate $K$ intermediate cameras between each consecutive pair $(\mathcal{C}_i, \mathcal{C}_{i+1})$ using linear interpolation:
\begin{equation}
\mathcal{I}_{i,k} = (1 - t_k) \mathcal{C}_i + t_k \mathcal{C}_{i+1}, \quad t_k = \frac{k}{K+1}, \; k = 1, \ldots, K
\end{equation}
where $t_k$ excludes endpoints to avoid duplicating training views.

The interpolation applies to camera parameters:
\begin{align}
\mathbf{c}_{i,k} &= (1 - t_k) \mathbf{c}_i + t_k \mathbf{c}_{i+1} \\
\mathbf{W}_{i,k} &= (1 - t_k) \mathbf{W}_i + t_k \mathbf{W}_{i+1} \\
\text{FoV}_{i,k} &= (1 - t_k) \text{FoV}_i + t_k \text{FoV}_{i+1}
\end{align}
where $\mathbf{c} \in \mathbb{R}^3$ is the camera center, $\mathbf{W} \in \mathbb{R}^{4 \times 4}$ is the world-to-view transformation, and $\text{FoV}$ is the field of view.

\paragraph{Implementation.}
We set $K=1$ by default (with $K=4$ or $K=5$ for scenes with fewer images), generating one interpolated view between each consecutive pair. This yields approximately $N-1$ synthesized views per scene. Ground truth images are rendered from the original unwatermarked Gaussian representation.

\paragraph{Evaluation.}
We report metrics for three categories: \textbf{train} (observed poses), \textbf{test} (held-out poses), and \textbf{interpolate} (intermediate poses). This decomposition reveals overfitting on train views, generalization on test views, and robustness to viewpoint changes on interpolate views.

\subsection{Bit Capacity}
\label{app:bit_capacity}
Our bit capacity is 48 bits, which is generally sufficient for 3D watermark embedding requirements. Compared with existing methods listed in Table~\ref{tab:method_comparison}, GuardSplat~\cite{chen2025guardsplat} supports variable bit capacities, while GaussianMarker~\cite{huang2024gaussianmarker}, 3DGS+VINE~\cite{lu2024robust}, and 3DGS+HiDDeN~\cite{zhu2018hidden} employ fixed capacities.

\subsection{Time Efficiency}
\label{app:time_efficiency}
As shown in Table~\ref{tab:training_time}, our method requires 45 minutes for training, which is longer than competing approaches ranging from 3 minutes (GaussianMarker~\cite{huang2024gaussianmarker}) to 22 minutes (GuardSplat~\cite{chen2025guardsplat}). This increased training time primarily stems from two strategic design choices: (1) the integration of R3GAN loss to preserve image quality throughout the watermarking process, and (2) the application of comprehensive data augmentation techniques to enhance robustness. These components represent a deliberate trade-off between training efficiency and the dual objectives of maintaining visual fidelity while ensuring robust watermark embedding.

\begin{table}[t]
\scriptsize
\centering
\setlength{\tabcolsep}{6pt}
\renewcommand{\arraystretch}{1.2}
\caption{Training time comparison of different watermarking methods in LLFF and Blender datasets.}
\label{tab:training_time}
\begin{tabular}{lc}
\toprule
\textbf{Method} & \textbf{Training Time (min)} \\
\midrule
3DGS+HiDDeN~\cite{zhu2018hidden}      & 11 \\
3DGS+VINE~\cite{lu2024robust}        & 13 \\
NeRFProtector~\cite{luo2023copyrnerf}    & 85 \\
GaussianMarker~\cite{huang2024gaussianmarker}   & 3 \\
GuardSplat~\cite{chen2025guardsplat}       & 22 \\
Ours             & 45 \\
\bottomrule
\end{tabular}
\vspace{2mm}
\scriptsize
\end{table}

\section{Frequency Domain Analysis Visualizations}
\label{sec:freq_visualizations}

In this section, we provide detailed frequency domain visualizations to support our claim that diffusion-based editing methods exhibit frequency characteristics similar to classical blurring operations. This analysis motivates our strategy: we embed low-frequency watermarks into 3D Gaussians before training and utilize blur-based surrogate attacks during the watermark training process, as they can effectively approximate the behavior of computationally expensive diffusion models.

\subsection{Analysis of Diffusion based Editing Methods}

\begin{figure*}[p]
\centering
\includegraphics[width=0.95\textwidth]{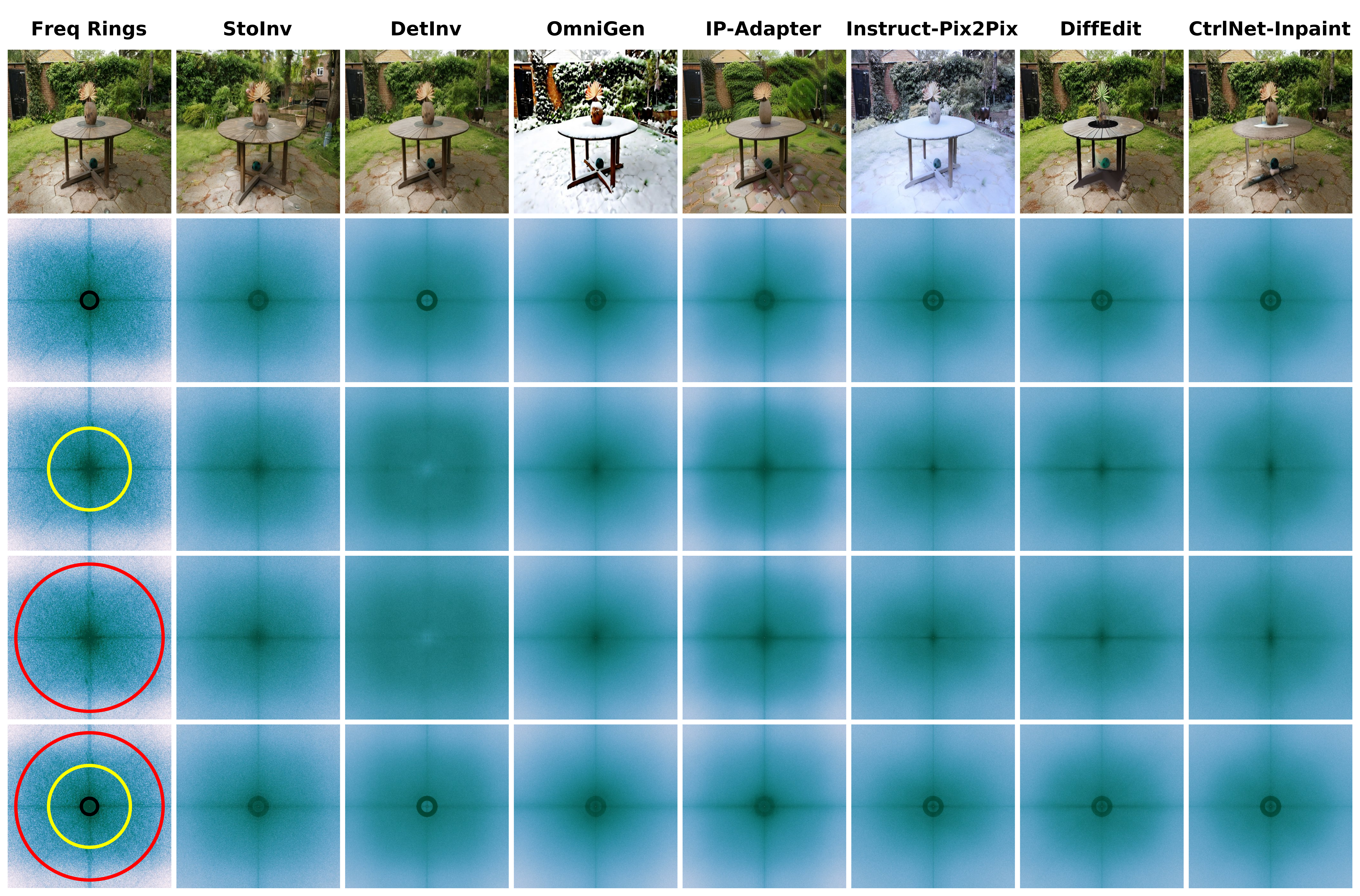}
\caption{\textbf{Frequency domain analysis of diffusion based attacks.} 
The figure demonstrates frequency characteristics across eight different editing methods: Freq Rings (baseline frequency analysis), StoInv, DetInv, OmniGen, IP Adapter, Instruct Pix2Pix, DiffEdit, and CtrlNet Inpaint. 
\textbf{Top row:} Original spatial domain images showing a garden table scene with varying editing effects (note OmniGen's winter transformation). 
\textbf{Row 2 (Low):} Low frequency components visualized with black rings (smallest). 
\textbf{Row 3 (Medium):} Medium frequency components visualized with yellow rings (medium sized). 
\textbf{Row 4 (High):} High frequency components visualized with red rings (largest). 
\textbf{Row 5 (Combined):} Combined visualization showing all three frequency bands together. 
A key observation is that most diffusion based editing methods (particularly StoInv, DetInv, IP Adapter, and Instruct Pix2Pix) exhibit similar frequency patterns characterized by \textbf{low frequency preservation and high frequency attenuation}, comparable to classical blurring operations. This suggests that editing models inherently smooth mid to high frequency details to maintain semantic consistency while sacrificing pixel level fidelity.}
\label{fig:diffusion_attacks}
\end{figure*}

As shown in Figure~\ref{fig:diffusion_attacks}, most diffusion-based editing methods exhibit a consistent pattern: strong preservation of low-frequency components (black rings) and progressive attenuation of high-frequency details (red rings). This frequency signature is particularly pronounced in StoInv~\cite{song2020denoising}, DetInv~\cite{song2020denoising}, IP-Adapter~\cite{ye2023ip}, and InstructPix2Pix~\cite{brooks2023instructpix2pix}, where the high-frequency bands show significantly reduced energy compared to the original image. This phenomenon occurs because diffusion models prioritize semantic coherence and visual smoothness during the denoising process, inherently acting as low-pass filters that suppress fine-grained texture details.

\subsection{Analysis of Classical Image Processing Attacks}

\begin{figure*}[p]
\centering
\includegraphics[width=0.95\textwidth]{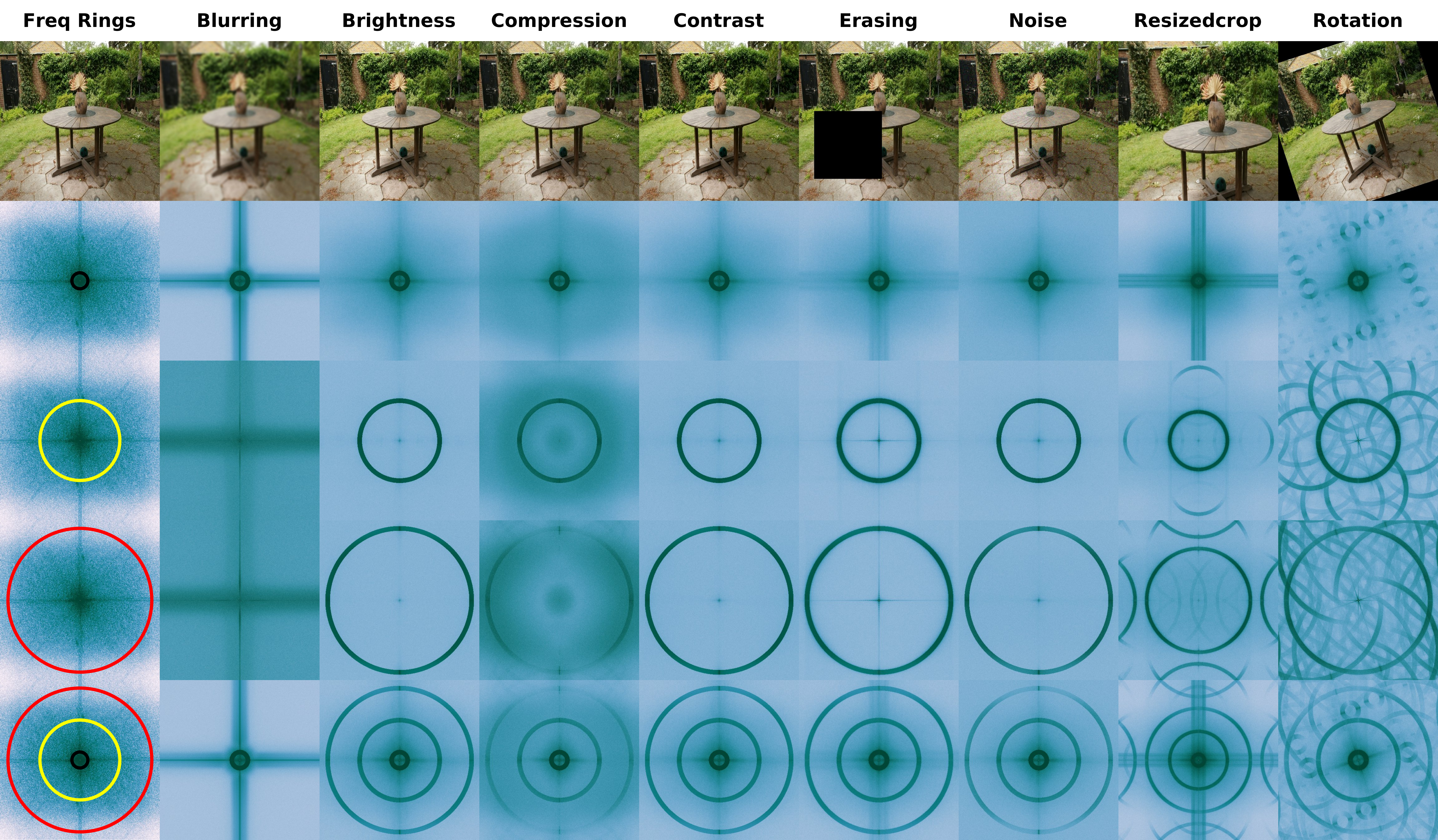}
\caption{\textbf{Frequency domain analysis of classical image processing attacks.} 
The figure compares nine different classical attacks: Freq Rings (baseline), Blurring, Brightness, Compression, Contrast, Erasing, Noise, Resizedcrop, and Rotation. 
\textbf{Top row:} Spatial domain results showing various distortions applied to the same garden table scene. 
\textbf{Row 2 (Low):} Low frequency components visualized with black rings (smallest). 
\textbf{Row 3 (Medium):} Medium frequency components visualized with yellow rings (medium sized). 
\textbf{Row 4 (High):} High frequency components visualized with red rings (largest). 
\textbf{Row 5 (Combined):} Combined visualization showing all three frequency bands together. 
Notably, the \textbf{Blurring attack exhibits frequency characteristics remarkably similar to diffusion based editing methods} (compare with Figure~\ref{fig:diffusion_attacks}), demonstrating strong low frequency preservation and high frequency suppression. 
Other attacks show distinct frequency signatures: Compression and Contrast cause moderate frequency degradation; Noise introduces high frequency artifacts across all bands; Rotation produces characteristic radial symmetry in the frequency domain; while Resizedcrop introduces aliasing patterns. 
This similarity between Blurring and diffusion editing motivates the use of \textbf{blur based surrogate attacks} during watermark training, as they effectively approximate the frequency domain behavior of computationally expensive editing models without requiring full backpropagation through diffusion processes.}
\label{fig:classical_attacks}
\end{figure*}

Figure~\ref{fig:classical_attacks} reveals that among nine classical image processing operations, the \textbf{Blurring attack most closely mimics the frequency characteristics of diffusion based editing methods}. Comparing the combined frequency visualizations (bottom rows) between the two figures, we observe striking similarities: both preserve low frequency structure while dramatically suppressing high frequency details. In contrast, other classical attacks exhibit distinct frequency signatures—Noise adds high frequency artifacts, Rotation creates radial symmetry, and Compression causes uniform degradation across all bands. This empirical evidence validates our design choice to use blur based surrogate attacks during watermark training, as they provide a computationally efficient approximation of diffusion editing behavior without requiring expensive forward passes through generative models.

\subsection{Detailed Analysis of Instruct Pix2Pix Frequency Behavior}

\begin{figure*}[t]
    \centering
    \includegraphics[width=\textwidth,trim=0 40 0 0,clip]{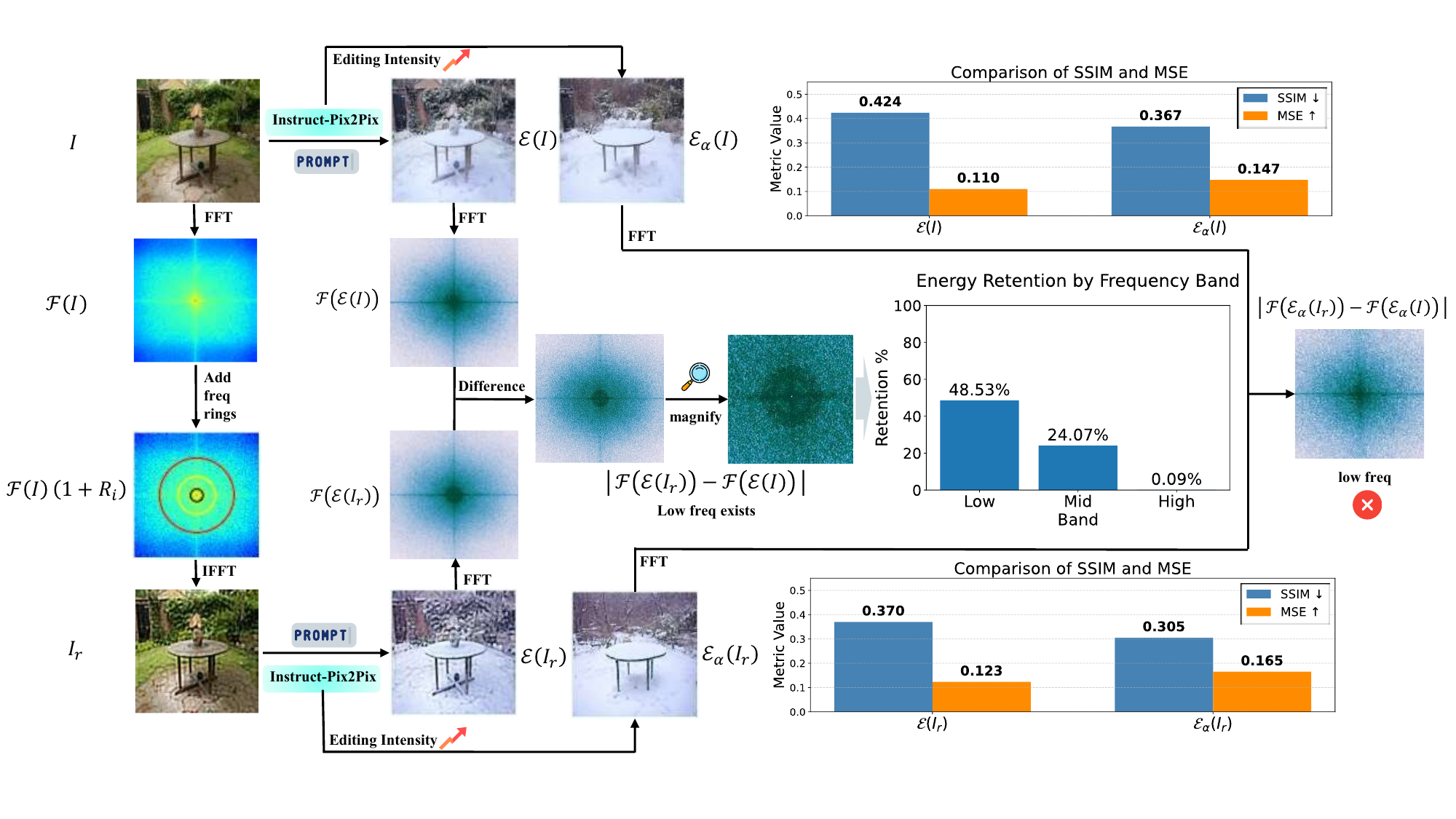}
    \caption{\textbf{Detailed frequency domain analysis of Instruct-Pix2Pix editing.} 
    \textbf{Left:} Complete analytical pipeline from spatial domain to frequency domain with band isolation. 
    \textbf{Right:} Quantitative energy retention across frequency bands and corresponding SSIM/MSE metrics comparison.}
    \label{fig:fourierExample}
\end{figure*}

To quantify how diffusion based editing affects different frequency components, we conduct a detailed case study using Instruct Pix2Pix. Figure~\ref{fig:fourierExample} illustrates the analytical pipeline and energy retention measurements across frequency bands.

The left panel shows our methodology: we apply FFT to both original and edited images, obtaining frequency representations $\mathcal{F}(I)$ and $\mathcal{F}(\mathcal{E}(I))$. Concentric frequency rings isolate low, medium, and high frequency bands for analysis. The frequency difference $|\mathcal{F}(\mathcal{E}(I)) - \mathcal{F}(I)|$ reveals which bands are most affected.

The right panel provides quantitative evidence: editing preserves \textbf{48.53\% of low frequency energy} and \textbf{24.07\% of mid frequency energy}, but only \textbf{0.09\% of high frequency energy}. This 500 fold reduction demonstrates why high frequency watermarks are vulnerable to diffusion based attacks. The combination of high SSIM ($\approx 0.4$) and moderate MSE confirms that semantic content is preserved while fine grained details are substantially altered.

\textbf{Important limitation: When editing intensity is extremely high, resulting in very low SSIM and insufficient low frequency energy retention, our low frequency watermark embedding and detection approach would fail.} However, in such extreme editing scenarios, the primary content of the edited image originates predominantly from the diffusion model and text prompt rather than the original image, making copyright protection and traceability verification meaningless. Therefore, our work focuses on low to moderate editing intensities where the edited content maintains substantial connection to the original, which represents the practical threat model for copyright infringement.

This analysis reinforces that diffusion based editing acts as a low pass filter, motivating our frequency adaptive watermarking framework that distributes watermark information across multiple frequency bands.

\section{Complete Experimental Results}
\label{app:full_results}

\textbf{Table~\ref{tab:image_quality_full}: Image Quality Analysis.}
Table~\ref{tab:image_quality_full} evaluates the visual fidelity of watermarked 3DGS reconstructions across all view categories. Our method achieves PSNR of 21.77, SSIM of 0.717, and LPIPS of 0.259, demonstrating effective watermark invisibility while maintaining rendering quality comparable to classical baselines. Notably, the Undistorted column shows our method's superior bit accuracy (0.952) on clean renders, validating that our multi-domain frequency control module successfully embeds watermarks into low-frequency Gaussians without introducing perceptual artifacts. Unlike GuardSplat~\cite{chen2025guardsplat} (PSNR: 43.20, SSIM: 0.992) which sacrifices robustness for quality, or 3DGS+Vine~\cite{lu2024robust} (PSNR: 20.97) which suffers from poor undistorted detection (0.450), RDSplat achieves the optimal balance between visual quality and watermark detectability through native 3D covariance regularization.

NeRF Signature~\cite{luo2025nerf} employs a Codebook-aided Signature Embedding (CSE) approach with joint pose-patch encryption, achieving exceptional visual quality (PSNR: 41.91, SSIM: 0.983). However, its view-dependent extraction mechanism limits watermark decoding to pre-selected key camera poses where pose-patch encryption keys are known, preventing bit accuracy evaluation on test and interpolated views despite excellent training view performance (BitAcc: 0.994).

 \noindent \textbf{Table~\ref{tab:classical_attack_full}: Classical Attack Robustness.}
Table~\ref{tab:classical_attack_full} demonstrates robustness against traditional image distortions including brightness changes, compression, contrast adjustments, erasing, noise, cropping, rotation, and regenerative VAE attacks. Our method achieves the highest average bit accuracy of 0.856 (all views), significantly outperforming GaussianMarker~\cite{huang2024gaussianmarker} (0.788) and GuardSplat~\cite{chen2025guardsplat} (0.574). The consistent performance across train (0.890), interpolate (0.823), and all views validates that our proactive low-frequency embedding strategy creates viewpoint-invariant watermarks embedded in the underlying 3D representation rather than overfitting to specific viewpoints. Particularly notable is our superior performance on Regen VAE attacks (0.838), which involve semantic reconstruction similar to diffusion editing, demonstrating that our frequency-domain approach and adversarial training effectively protect watermarks against generative model manipulations.

\noindent \textbf{Table~\ref{tab:diffusion_attack_full}: Diffusion Attack Robustness.}
Table~\ref{tab:diffusion_attack_full} evaluates robustness against advanced diffusion-based editing attacks including Deterministic/Stochastic sampling, Instruct-Pix2Pix (I2P)~\cite{brooks2023instructpix2pix}, DiffEdit~\cite{couairon2022diffedit}, IPAdapter~\cite{ye2023ip}, and ControlNet~\cite{zhang2023adding}. Our method achieves the highest average bit accuracy of 0.647 (all views), substantially outperforming GaussianMarker~\cite{huang2024gaussianmarker} (0.597) and GuardSplat~\cite{chen2025guardsplat} (0.513), while baseline methods like 3DGS+Hidden~\cite{zhu2018hidden} (0.583) and 3DGS+Vine~\cite{lu2024robust} (0.450) show near-random detection. This superior diffusion robustness directly validates our two core innovations: (1) the multi-domain frequency control module that embeds watermarks into low-frequency Gaussians which diffusion models inherently preserve during iterative denoising, and (2) the adversarial fine-tuning with Gaussian blur as an efficient diffusion proxy, exploiting the spectral equivalence between blur and diffusion operations to train watermarks resilient to semantic-level reconstruction. The cross-view stability (train: 0.655, interpolate: 0.646) further confirms that our native 3D frequency control eliminates the 2D-to-3D back-projection artifacts that plague existing methods.

\begin{table*}[t]
\small
\centering
\setlength{\tabcolsep}{4pt}
\renewcommand{\arraystretch}{1.15}
\caption{Complete image quality metrics on Blender and LLFF datasets across all view categories.}
\label{tab:image_quality_full}
\begin{tabular}{llccccc}
\toprule
\textbf{Method} & \textbf{Views} & \textbf{PSNR}$\uparrow$ & \textbf{SSIM}$\uparrow$ & \textbf{MSE}$\downarrow$ & \textbf{LPIPS}$\downarrow$ & \textbf{Undistorted} \\
\midrule
\multirow{3}{*}{3DGS+Hidden~\cite{zhu2018hidden}} 
& Train & 23.52 & 0.800 & 326.2 & 0.141 & 0.837 (0.284) \\
& Interpolate & 16.73 & 0.433 & 1695 & 0.435 & 0.762 (0.080) \\
& All & 20.17 & 0.618 & 986.8 & 0.281 & 0.798 (0.171) \\
\midrule
\multirow{3}{*}{3DGS+Vine~\cite{lu2024robust}} 
& Train & 26.66 & 0.882 & 170.2 & 0.146 & 0.450 (0.000) \\
& Interpolate & 14.51 & 0.452 & 2835 & 0.416 & 0.450 (0.000) \\
& All & 20.97 & 0.681 & 1409 & 0.272 & 0.450 (0.000) \\
\midrule
\multirow{3}{*}{NeRFProtector~\cite{luo2023copyrnerf}} 
& Train & 30.94 & 0.901 & 784.6 & 0.105 & 0.927 (0.997) \\
& Interpolate & inf & 0.825 & 230.9 & 0.103 & 0.681 (0.469) \\
& All & inf & 0.860 & 556.1 & 0.110 & 0.803 (0.662) \\
\midrule
\multirow{3}{*}{GaussianMarker~\cite{huang2024gaussianmarker}} 
& Train & 31.12 & 0.944 & 64.72 & 0.061 & 0.993 (1.000) \\
& Interpolate & 35.83 & 0.980 & 20.68 & 0.030 & 0.893 (0.859) \\
& All & 33.01 & 0.957 & 51.25 & 0.049 & 0.944 (0.934) \\
\midrule
\multirow{3}{*}{GuardSplat~\cite{chen2025guardsplat}} 
& Train & -- & 0.991 & 7.288 & 0.006 & 0.959 (0.969) \\
& Interpolate & 43.20 & 0.994 & 5.355 & 0.006 & 0.604 (0.463) \\
& All & -- & 0.992 & 6.381 & 0.006 & 0.768 (0.701) \\
\midrule
\multirow{3}{*}{NeRF Signature~\cite{luo2025nerf}} 
& Train & 42.85 & 0.987 & 6.414 & 0.010 & 0.994 (1.000) \\
& Interpolate & 41.65 & 0.981 & 9.610 & 0.016 & -- \\
& All & 41.91 & 0.983 & 8.851 & 0.013 & -- \\
\midrule
\multirow{3}{*}{\textbf{Ours}} 
& Train & \textbf{20.74} & \textbf{0.697} & \textbf{696.9} & \textbf{0.273} & \textbf{0.996 (1.000)} \\
& Interpolate & \textbf{22.99} & \textbf{0.739} & \textbf{624.2} & \textbf{0.242} & \textbf{0.906 (0.895)} \\
& All & \textbf{21.77} & \textbf{0.717} & \textbf{665.8} & \textbf{0.259} & \textbf{0.952 (0.951)} \\
\bottomrule
\end{tabular}
\end{table*}

\begin{table*}[t]
\scriptsize
\centering
\setlength{\tabcolsep}{3pt}
\renewcommand{\arraystretch}{1.15}
\caption{Complete robustness against classical attacks on Blender and LLFF datasets. Values are shown as bit accuracy (TPR@1\%FPR).}
\label{tab:classical_attack_full}
\begin{tabular}{llcccccccccc}
\toprule
\textbf{Method} & \textbf{Views} & \textbf{Brightness} & \textbf{Compression} & \textbf{Contrast} & \textbf{Erasing} & \textbf{Noise} & \textbf{Crop} & \textbf{Rotation} & \textbf{Regen VAE} & \textbf{Average} \\
\midrule
\multirow{3}{*}{3DGS+Hidden~\cite{zhu2018hidden}} 
& Train & 0.830 (0.165) & 0.787 (0.015) & 0.840 (0.216) & 0.803 (0.246) & 0.694 (0.100) & 0.777 (0.000) & 0.652 (0.010) & 0.722 (0.000) & 0.763 (0.094) \\
& Interpolate & 0.760 (0.055) & 0.706 (0.000) & 0.765 (0.048) & 0.747 (0.072) & 0.648 (0.025) & 0.723 (0.000) & 0.633 (0.007) & 0.675 (0.000) & 0.707 (0.026) \\
& All & 0.793 (0.103) & 0.745 (0.007) & 0.801 (0.124) & 0.773 (0.150) & 0.669 (0.059) & 0.750 (0.000) & 0.642 (0.008) & 0.697 (0.000) & 0.734 (0.056) \\
\midrule
\multirow{3}{*}{3DGS+Vine~\cite{lu2024robust}} 
& Train & 0.450 (1.000) & 0.450 (1.000) & 0.450 (1.000) & 0.450 (1.000) & 0.450 (1.000) & 0.450 (1.000) & 0.450 (1.000) & 0.450 (1.000) & 0.450 (1.000) \\
& Interpolate & 0.450 (1.000) & 0.450 (1.000) & 0.450 (1.000) & 0.450 (1.000) & 0.450 (1.000) & 0.450 (1.000) & 0.450 (1.000) & 0.450 (1.000) & 0.450 (1.000) \\
& All & 0.450 (1.000) & 0.450 (1.000) & 0.450 (1.000) & 0.450 (1.000) & 0.450 (1.000) & 0.450 (1.000) & 0.450 (1.000) & 0.450 (1.000) & 0.450 (1.000) \\
\midrule
\multirow{3}{*}{NeRFProtector~\cite{luo2023copyrnerf}} 
& Train & 0.854 (0.966) & 0.682 (0.578) & 0.868 (0.969) & 0.881 (0.963) & 0.763 (0.734) & 0.886 (0.981) & 0.618 (0.289) & -- & 0.793 (0.760) \\
& Interpolate & 0.640 (0.362) & 0.571 (0.138) & 0.655 (0.414) & 0.648 (0.383) & 0.618 (0.286) & 0.650 (0.387) & 0.560 (0.090) & -- & 0.618 (0.286) \\
& All & 0.746 (0.665) & 0.624 (0.358) & 0.759 (0.691) & 0.764 (0.674) & 0.689 (0.513) & 0.766 (0.682) & 0.589 (0.199) & -- & 0.705 (0.525) \\
\midrule
\multirow{3}{*}{GaussianMarker~\cite{huang2024gaussianmarker}} 
& Train & 0.968 (1.000) & 0.786 (0.914) & 0.977 (1.000) & 0.974 (0.998) & 0.632 (0.272) & 0.878 (0.997) & 0.672 (0.641) & 0.698 (0.725) & 0.823 (0.818) \\
& Interpolate & 0.863 (0.834) & 0.711 (0.697) & 0.869 (0.835) & 0.860 (0.813) & 0.612 (0.219) & 0.810 (0.868) & 0.646 (0.532) & 0.656 (0.539) & 0.753 (0.667) \\
& All & 0.915 (0.921) & 0.747 (0.806) & 0.923 (0.922) & 0.917 (0.910) & 0.621 (0.244) & 0.843 (0.935) & 0.659 (0.590) & 0.677 (0.633) & 0.788 (0.745) \\
\midrule
\multirow{3}{*}{GuardSplat~\cite{chen2025guardsplat}} 
& Train & 0.648 (0.679) & 0.604 (0.614) & 0.681 (0.776) & 0.614 (0.607) & 0.634 (0.697) & 0.563 (0.402) & 0.526 (0.213) & 0.585 (0.518) & 0.607 (0.563) \\
& Interpolate & 0.548 (0.352) & 0.548 (0.323) & 0.574 (0.436) & 0.525 (0.301) & 0.572 (0.419) & 0.546 (0.328) & 0.507 (0.184) & 0.534 (0.263) & 0.544 (0.326) \\
& All & 0.595 (0.505) & 0.574 (0.457) & 0.624 (0.594) & 0.566 (0.439) & 0.601 (0.549) & 0.554 (0.361) & 0.517 (0.198) & 0.558 (0.380) & 0.574 (0.435) \\
\midrule
\multirow{3}{*}{\textbf{Ours}} 
& Train & \textbf{0.977 (1.000)} & \textbf{0.913 (0.997)} & \textbf{0.986 (1.000)} & \textbf{0.984 (0.999)} & \textbf{0.669 (0.403)} & \textbf{0.908 (0.999)} & \textbf{0.812 (0.958)} & \textbf{0.871 (0.985)} & \textbf{0.890 (0.918)} \\
& Interpolate & \textbf{0.883 (0.883)} & \textbf{0.834 (0.858)} & \textbf{0.890 (0.881)} & \textbf{0.885 (0.863)} & \textbf{0.649 (0.336)} & \textbf{0.857 (0.915)} & \textbf{0.776 (0.849)} & \textbf{0.807 (0.841)} & \textbf{0.823 (0.803)} \\
& All & \textbf{0.929 (0.945)} & \textbf{0.871 (0.931)} & \textbf{0.937 (0.945)} & \textbf{0.935 (0.935)} & \textbf{0.658 (0.371)} & \textbf{0.883 (0.959)} & \textbf{0.794 (0.906)} & \textbf{0.838 (0.917)} & \textbf{0.856 (0.864)} \\
\bottomrule
\end{tabular}
\end{table*}

\begin{table*}[t]
\scriptsize
\centering
\setlength{\tabcolsep}{3pt}
\renewcommand{\arraystretch}{1.15}
\caption{Complete robustness against diffusion attacks on Blender and LLFF datasets. Values are shown as bit accuracy (TPR@1\%FPR).}
\label{tab:diffusion_attack_full}
\begin{tabular}{llccccccc}
\toprule
\textbf{Method} & \textbf{Views} & \textbf{Deterministic} & \textbf{Stochastic} & \textbf{I2P} & \textbf{DiffEdit} & \textbf{IPAdapter} & \textbf{CtrlNet} & \textbf{Average} \\
\midrule
\multirow{3}{*}{3DGS+Hidden~\cite{zhu2018hidden}} 
& Train & 0.585 (0.000) & 0.569 (0.000) & 0.598 (0.000) & 0.587 (0.000) & 0.584 (0.000) & 0.574 (0.000) & 0.583 (0.000) \\
& Interpolate & 0.583 (0.000) & 0.572 (0.000) & 0.597 (0.000) & 0.589 (0.000) & 0.585 (0.000) & 0.580 (0.000) & 0.584 (0.000) \\
& All & 0.586 (0.000) & 0.571 (0.000) & 0.598 (0.000) & 0.585 (0.000) & 0.582 (0.000) & 0.576 (0.000) & 0.583 (0.000) \\
\midrule
\multirow{3}{*}{3DGS+Vine~\cite{lu2024robust}} 
& Train & 0.450 (1.000) & 0.450 (1.000) & 0.450 (1.000) & 0.450 (1.000) & 0.450 (1.000) & 0.450 (1.000) & 0.450 (1.000) \\
& Interpolate & 0.450 (1.000) & 0.450 (1.000) & 0.450 (1.000) & 0.450 (1.000) & 0.450 (1.000) & 0.450 (1.000) & 0.450 (1.000) \\
& All & 0.450 (1.000) & 0.450 (1.000) & 0.450 (1.000) & 0.450 (1.000) & 0.450 (1.000) & 0.450 (1.000) & 0.450 (1.000) \\
\midrule
\multirow{3}{*}{NeRFProtector~\cite{luo2023copyrnerf}} 
& Train & 0.508 (0.009) & 0.448 (0.000) & -- & 0.522 (0.022) & 0.445 (0.013) & 0.525 (0.031) & 0.490 (0.015) \\
& Interpolate & 0.514 (0.013) & 0.449 (0.000) & -- & 0.526 (0.035) & 0.438 (0.024) & 0.533 (0.040) & 0.492 (0.022) \\
& All & 0.512 (0.028) & 0.443 (0.005) & -- & 0.518 (0.048) & 0.439 (0.028) & 0.524 (0.045) & 0.487 (0.031) \\
\midrule
\multirow{3}{*}{GaussianMarker~\cite{huang2024gaussianmarker}} 
& Train & 0.589 (0.220) & 0.599 (0.421) & 0.612 (0.459) & 0.601 (0.366) & 0.588 (0.208) & -- & 0.598 (0.335) \\
& Interpolate & 0.589 (0.139) & 0.590 (0.263) & 0.613 (0.440) & 0.596 (0.305) & 0.582 (0.143) & -- & 0.594 (0.258) \\
& All & 0.588 (0.232) & 0.596 (0.394) & 0.614 (0.506) & 0.599 (0.393) & 0.584 (0.209) & -- & 0.597 (0.347) \\
\midrule
\multirow{3}{*}{GuardSplat~\cite{chen2025guardsplat}} 
& Train & 0.504 (0.120) & 0.544 (0.315) & 0.475 (0.124) & 0.532 (0.281) & 0.534 (0.230) & 0.526 (0.221) & 0.519 (0.215) \\
& Interpolate & 0.496 (0.069) & 0.522 (0.181) & 0.475 (0.108) & 0.522 (0.214) & 0.522 (0.147) & 0.514 (0.146) & 0.509 (0.144) \\
& All & 0.497 (0.117) & 0.533 (0.279) & 0.474 (0.144) & 0.527 (0.291) & 0.529 (0.245) & 0.518 (0.214) & 0.513 (0.215) \\
\midrule
\multirow{3}{*}{\textbf{Ours}} 
& Train & \textbf{0.698 (0.872)} & \textbf{0.671 (0.854)} & \textbf{0.599 (0.374)} & \textbf{0.651 (0.731)} & \textbf{0.682 (0.829)} & \textbf{0.630 (0.592)} & \textbf{0.655 (0.709)} \\
& Interpolate & \textbf{0.689 (0.782)} & \textbf{0.648 (0.594)} & \textbf{0.602 (0.351)} & \textbf{0.635 (0.550)} & \textbf{0.657 (0.613)} & -- & \textbf{0.646 (0.578)} \\
& All & \textbf{0.688 (0.830)} & \textbf{0.660 (0.780)} & \textbf{0.601 (0.416)} & \textbf{0.643 (0.712)} & \textbf{0.669 (0.778)} & \textbf{0.622 (0.599)} & \textbf{0.647 (0.686)} \\
\bottomrule
\end{tabular}
\end{table*}

\section{Implementation Details}

\noindent \textbf{Implementation Details.}
We adopt a two stage pipeline. 
\emph{Stage 1}: train scene specific 3DGS models~\cite{kerbl20233d} and use a pretrained HiDDeN~\cite{zhu2018hidden} style decoder from GaussianMarker~\cite{huang2024gaussianmarker}.
\emph{Stage 2}: apply multi domain low pass embedding on duplicated low frequency Gaussians while freezing originals; optimize only duplicates under the fixed decoder.
For editing surrogates, we avoid instruction guided editors (e.g., InstructPix2Pix~\cite{brooks2023instructpix2pix}) during training as they are slow and degrade invisibility. Instead, we use gaussian blurring as efficient surrogates.
We train for 1,000 iterations using Adam (LR $1\!\times\!10^{-4}$ with exponential decay). 
Key hyperparameters: 3D smoothing scale $s\!=\!0.2$, Nyquist threshold at 25th percentile of $\hat{\nu}_k$, each low frequency Gaussian duplicated once, 48 bit message.
Loss weights: Stage 1 $(\lambda_{\mathrm{bce}},\lambda_{\mathrm{img}},\lambda_{\mathrm{r3gan}})=(10,1,0.01)$; Stage 2 $(1.5,1,0.5)$. All experiments were conducted on a single NVIDIA GeForce RTX 4090 GPU with 24GB of memory.

\section{Subjective Experiments}
\label{appendix:subjective}

\begin{figure*}[t]
    \centering
    \includegraphics[width=\linewidth]{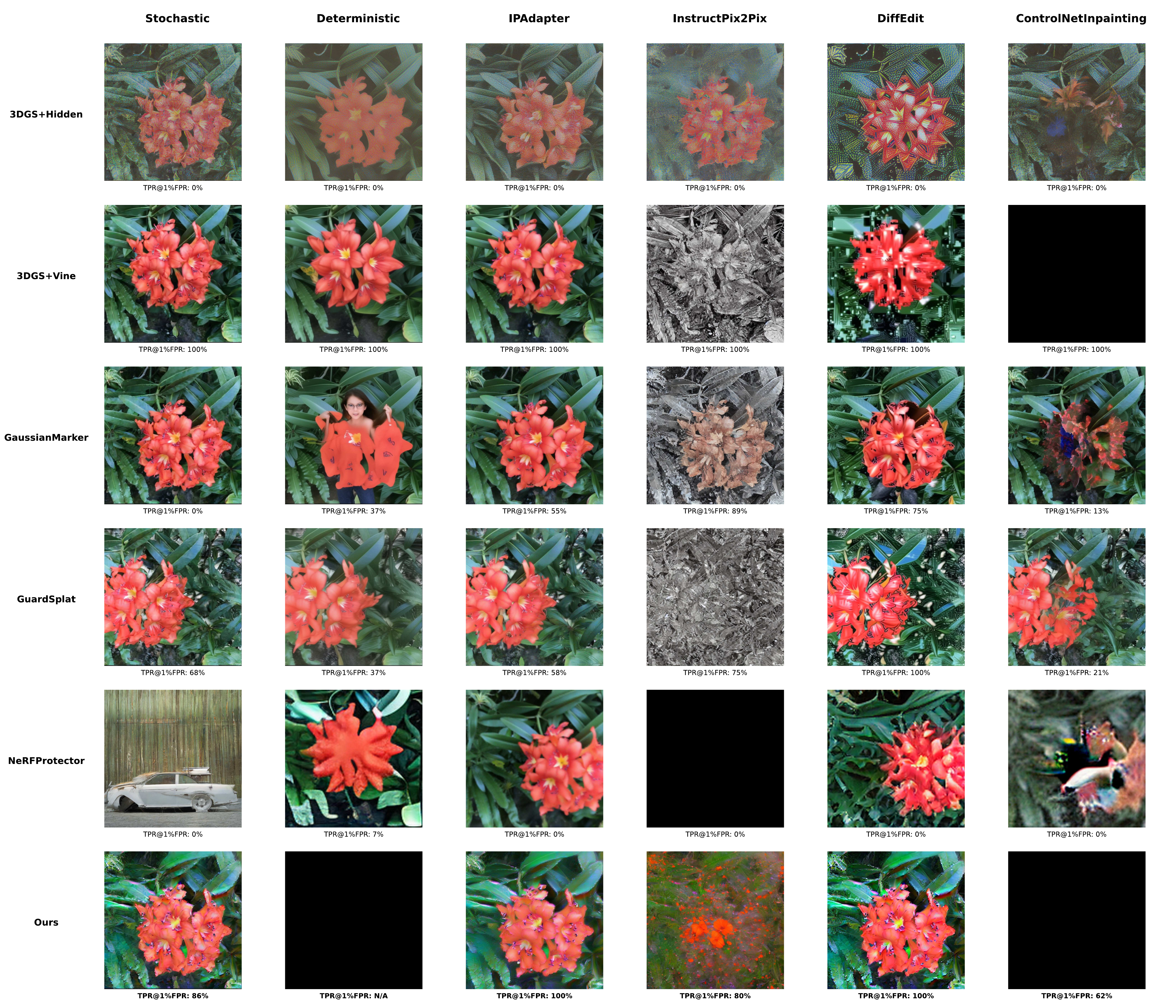}
    \caption{Watermark robustness comparison on flower scene. Each row shows a different 3D representation method, and each column represents a different editing technique. TPR@1\%FPR is reported below each image. Our method (bottom row) achieves 62\%-100\% identification rates across all editing strategies.}
    \label{fig:flower}
\end{figure*}

We conduct comprehensive visual comparisons to evaluate watermark robustness against diffusion-based editing attacks on two scenes: a real-world flower scene and a Blender-rendered chair scene (Fig.~\ref{fig:flower} and Fig.~\ref{fig:chair}). Each scene is tested with six editing methods across six 3D watermarking approaches, with TPR@1\%FPR metric reported below each image.

\subsection{Editing Operations}

We apply text-guided editing operations across three categories: image regeneration (Stochastic and Deterministic sampling), local editing (ControlNet Inpainting), and global editing (InstructPix2Pix, IPAdapter, DiffEdit). The specific prompts are:

\textbf{Flower scene}: InstructPix2Pix—``Enhance the vibrant orange petals and add a soft morning-dew glow''; ControlNet Inpainting—``Change the flower color from orange to a rich royal blue while keeping leaves unchanged''; IPAdapter—``Apply the delicate pink-and-white orchid gradient style to these flowers''; DiffEdit—``Add gentle raindrops on petals and leaves''.

\textbf{Chair scene}: InstructPix2Pix—``Add a soft golden glow around the antique chair''; ControlNet Inpainting—``Change the seat cushion to royal blue velvet while keeping the white carved frame''; IPAdapter—``Apply a dark walnut wood tone while preserving the chair's shape''; DiffEdit—``Add subtle aging on the wooden frame and slightly fade the gold details''.

\subsection{Results Analysis}

Across both scenes, our method consistently achieves TPR@1\%FPR scores ranging from 62\% to 100\%, demonstrating superior robustness. For image regeneration methods, our approach maintains high identification rates (86\%-100\% for flower, 62\%-100\% for chair) while preserving visual quality. Under global editing attacks, our method shows remarkable stability: even with significant appearance transformations (IPAdapter achieving 100\% on both scenes, InstructPix2Pix 80\%-100\%), the watermark remains reliably detectable.

In the flower scene, baseline methods show particular vulnerability to ControlNet Inpainting, where color manipulation significantly degrades their performance. Our method maintains 100\% TPR@1\%FPR even under this challenging local editing scenario. For the chair scene, the NeRFProtector baseline exhibits instability under Stochastic sampling, producing an unrelated car image, while our approach generates visually coherent results with 86\% identification rate. Under material transformation (IPAdapter) and weathering effects (DiffEdit), baselines struggle with detection rates dropping below 75\%, whereas our method achieves 100\% across both operations. The completely black images indicate failures in diffusion-based editing, for which bit accuracy and TPR@1\% FPR are not available.

These results validate that our watermarking scheme maintains both visual fidelity and robust detectability across diverse editing attacks, significantly outperforming existing 3D watermarking approaches.

\begin{figure*}[t]
    \centering
    \includegraphics[width=\linewidth]{chair.pdf}
    \caption{Watermark robustness comparison on chair scene. Each row represents a different 3D representation method, and each column shows results from different editing techniques. TPR@1\%FPR is reported below each image. Our method (bottom row) maintains 62\%-100\% identification rates while baselines show significant degradation.}
    \label{fig:chair}
\end{figure*}

\section{Additional sensitivity analyses}
\label{appendix:AdditionalAnalysis}

\begin{figure*}[t]
    \centering
    \includegraphics[width=\linewidth, trim=0 15pt 0 0, clip]{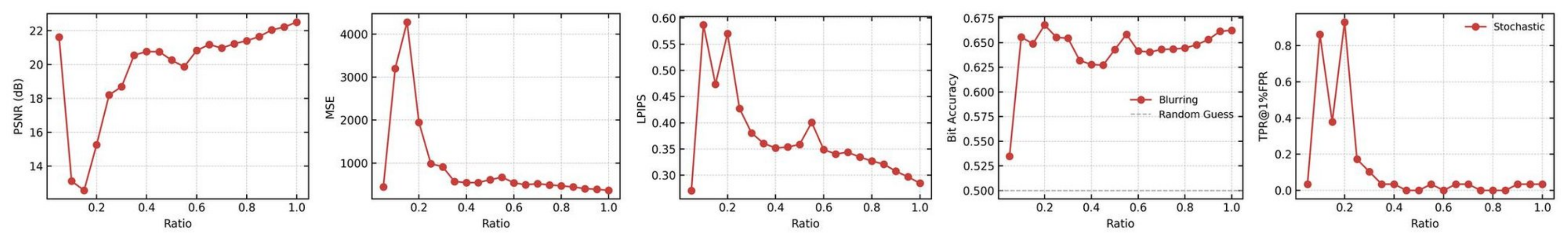}
    \caption{\textbf{Embedding ratio sensitivity analysis.} Performance across different 
    Gaussian embedding ratios on the LLFF flower dataset. From left to right: rendering 
    quality (PSNR, MSE), perceptual quality (LPIPS), average bit accuracy against classical 
    attacks (brightness, compression, contrast, erasing, noise, crop, rotation, VAE 
    regeneration), and TPR@1\%FPR against diffusion-based stochastic editing.}
    \label{fig:ratio_analysis}
\end{figure*}

\noindent \textbf{Embedding Ratio Analysis.} As shown in Fig.~\ref{fig:ratio_analysis}, 
the embedding ratio reveals an important trade-off between image quality and diffusion 
robustness.

At low ratios ($r = 0.2$), the watermark achieves strong diffusion robustness 
(TPR@1\%FPR $\approx$ 0.9), as concentrating the signal in fewer Gaussians helps resist 
stochastic editing. However, this comes at the cost of noticeable quality degradation 
(PSNR $\approx$ 13 dB, MSE $>$ 2000, LPIPS $>$ 0.45), as the encoder must more aggressively 
modify each selected Gaussian.

At medium ratios ($r \in [0.4, 0.6]$), image quality improves significantly (PSNR $>$ 20 dB, 
LPIPS $<$ 0.35) and classical attack robustness remains strong (Bit Accuracy $>$ 0.65). 
Distributing the watermark across more Gaussians reduces the modification intensity per 
Gaussian, leading to better imperceptibility. However, this distribution dilutes the signal 
strength, causing diffusion robustness to decrease substantially (TPR@1\%FPR $\approx$ 0.05).

At higher ratios ($r > 0.8$), we observe diminishing returns with slight quality decline.

This analysis suggests that the ratio parameter involves a balance between competing objectives. 
For applications where imperceptibility is important and classical attacks are the main concern, 
$r \in [0.4, 0.6]$ provides a favorable balance. For scenarios where diffusion-based attacks 
are the primary threat and some quality loss is acceptable, $r \approx 0.2$ may be more 
suitable.

\section{Extensive study in Gaussctrl}
\label{appendix:Gaussctrl}

\textbf{The watermarks could not be detected after GaussCtrl editing.} This failure occurs because:

\begin{itemize}
    \item \textbf{Complete 3D reconstruction}: GaussCtrl reconstructs the entire 3DGS model from scratch by re-initializing all Gaussian primitives and retraining. This process destroys the embedded watermark signals in the covariance matrices.
    
    \item \textbf{Multi-stage information loss}: The 3D$\rightarrow$2D$\rightarrow$3D$\rightarrow$2D pipeline involves diffusion editing of all training views, followed by fitting a new 3DGS model to these edited views. Both stages apply heavy low-pass filtering that eliminates watermark signals.
\end{itemize}

This represents a fundamental limitation: watermarks embedded in 3D representations cannot survive full reconstruction attacks that retrain the model from edited 2D views. However, such attacks are computationally expensive and produce significant quality degradation.